\documentclass[10pt]{article}

\usepackage[nonatbib, final]{neurips_2020}
\usepackage[square,numbers]{natbib}
\usepackage[utf8]{inputenc} 
\usepackage[T1]{fontenc}    
\usepackage{hyperref}       
\usepackage{url}            
\usepackage{booktabs}       
\usepackage{amsfonts}       
\usepackage{nicefrac}       
\usepackage{microtype}      
\usepackage{graphicx}
\usepackage{url}            
\usepackage{booktabs}       
\usepackage{amsfonts}       
\usepackage{nicefrac}       
\usepackage{hyperref}
\usepackage{multirow}
\usepackage{xspace}
\usepackage{amsmath}
\usepackage{amsfonts}
\usepackage{amssymb,algorithm}
\usepackage{algpseudocode}
\usepackage{textcomp}
\usepackage{enumitem}
\usepackage{subcaption}
\usepackage{float}
\usepackage{wrapfig}
\usepackage{enumitem}
\newcommand{\rpool}{\ensuremath{{\rm RNNPool}}\xspace}
\newcommand{\rpoollayer}{\ensuremath{{\rm RNNPoolLayer}}\xspace}
\newcommand{\rpooln}{\ensuremath{{\rm RNNPool}}}
\newcommand{\alg}{\rpool}

\newtheorem{proposition}{Proposition}
\newtheorem{claim}{Claim}

\hypersetup{
    colorlinks=true,
    linkcolor=blue,
    filecolor=magenta,      
    urlcolor=cyan,
}
\begin{document}


\title{\rpool: Efficient Non-linear Pooling for RAM Constrained Inference}
\author{
Oindrila Saha$^\dagger$, Aditya Kusupati$^\ddagger$, \\\textbf{Harsha Vardhan Simhadri$^\dagger$, Manik Varma$^\dagger$ and Prateek Jain$^\dagger$}\\
$^\dagger$Microsoft Research India, 
$^\ddagger$University of Washington\\
\texttt{\{t-oisaha,harshasi,manik,prajain\}@microsoft.com}, \texttt{kusupati@cs.washington.edu}
}

\maketitle
\begin{abstract}
Standard Convolutional Neural Networks (CNNs) designed for computer
vision tasks tend to have large intermediate activation maps. These
require large working memory and are thus unsuitable for deployment on
resource-constrained devices typically used for inference on the
edge. Aggressively downsampling the images via pooling or strided
convolutions can address the problem but leads to a significant
decrease in accuracy due to gross aggregation of the feature map by
standard pooling operators. In this paper, we introduce \rpool, a
novel pooling operator based on Recurrent Neural Networks (RNNs), that
efficiently aggregates features over large patches of an image and
rapidly downsamples activation maps.  Empirical evaluation indicates
that an \rpool layer can effectively replace multiple blocks in a
variety of architectures such as
MobileNets,
DenseNet when applied to standard vision
tasks like image classification and face detection. That is, \rpool
can significantly decrease computational complexity and peak memory
usage for inference while retaining comparable accuracy. We use \rpool
with the standard S3FD~\citep{zhang2017s3fd} architecture to construct
a face detection method that achieves state-of-the-art MAP for tiny
ARM Cortex-M4 class microcontrollers with under $256$ KB of
RAM. Code is released at~\url{https://github.com/Microsoft/EdgeML}.
\end{abstract}

\section{Introduction}
\label{sec:intro}

Convolutional Neural Networks (CNNs) have become ubiquitous for
computer vision tasks such as image classification and face
detection. Steady progress has led to new CNN architectures that are
increasingly accurate, but also require larger memory and more
computation for inference. The increased inference complexity renders
these models unsuitable for resource-constrained processors that are
commonplace on the edge in IoT systems and battery-powered and
privacy-centric devices.

To reduce inference complexity, several techniques like
quantization~\citep{wang2019haq},
sparsification~\citep{gale2019state,kusupati2020soft}, cheaper CNN
blocks~\citep{sandler2018mobilenetv2,iandola2016squeezenet}, or neural
architecture search~\citep{tan2019efficientnet} have been proposed to
train CNN models with lower inference cost and model size while
retaining accuracy. However, these models still require large working
memory for inference. Memory tends to be the most constrained resource
on low power devices as it occupies a large fraction of the device die and
has high sustained power requirement~\citep{kim2017evaluating}. Most
low power ARM Cortex-M* microcontrollers have less than 256 KB~RAM.

Typical CNNs have large intermediate activation maps, as well as many
convolution layers, which put together require large amount of RAM for
inference (see Proposition~\ref{prop:mem}).  A standard approach to
reducing working memory is to use pooling operators or strided
convolution to bring down size of the activation map. In fact,
standard CNNs have multiple such layers. However, such pooling
operators aggregate the underlying activation map in a simplistic
manner, which can lead to a significant loss of accuracy. As a result,
their use is limited to small receptive fields, typically no larger
than $3\times 3$, and they can not be used to aggressively reduce the
activation map by aggregating larger receptive fields.

\begin{wrapfigure}{r}{0.5\columnwidth}
\vspace{-9pt}
 \centering
 \includegraphics[width=0.5\columnwidth]{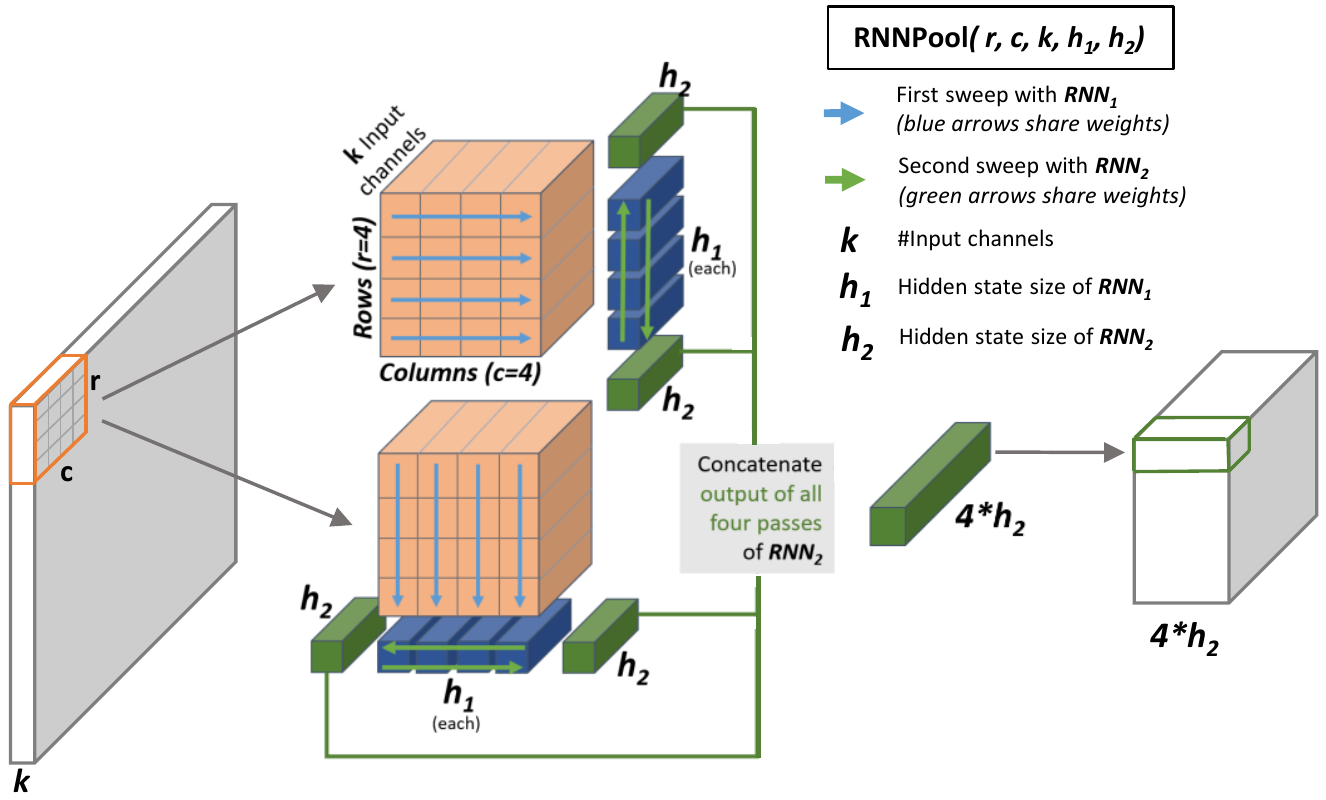}
 \vspace{-13pt}
 \caption{The \rpool operator applied to patches of size $r \times c$
   with stride $s$. It summarizes the patch with two RNNs into a
   vector of size $4h_2$.}
\label{fig:rnnpoolillus}	
\vspace{-10pt}
\end{wrapfigure}
 In this paper, we propose a
novel pooling operator \rpool that uses Recurrent Neural Networks
(RNNs) to perform a more refined aggregation over a large receptive field of
the activation map without compromising on accuracy. \rpool can be applied
to any tensor structured problem, but we focus on 2D images for ease
of exposition. For images, \rpool uses RNNs to aggregate information
along rows \& columns in a given patch. \rpool has three parameters --
patch size or receptive field, stride, and output dimension -- to
control its expressiveness and ability to downsample. The \rpool
operator matches standard pooling operators syntactically, so can be used to replace
them in convolutional networks.

\rpool allows rapid down-sampling of images and activation maps,
eliminating the need for many memory-intensive intermediate
layers. \rpool is most effective when used to replace multiple CNN
blocks in the initial stages of the network where the activation map
sizes are large, and hence, require the most memory and compute.
There, a single layer of \rpool can down-sample by a factor of $4$ or
$8$. For example, \rpool applied to a $640\times640\times3$ image
with patch-size 16, stride 8, and 32 output channels results in a
$80\times80\times32$ activation map, which can be stored in about
$200$ KB, and can be computed one patch at a time without significant
memory cost.  Replacing a few blocks using \rpool reduces peak memory
requirement significantly for typical CNN architectures without much
loss of accuracy.

Our experiments demonstrate that \rpool can be used as an effective
replacement for multi-layered, expensive CNN blocks in a variety of
architectures such as MobileNets, DenseNets, S3FD, and for varied
tasks such as image classification and face detection.  For example,
in a 10-class image classification task, \rpooln+MobileNetV2 reduces
the peak memory requirement of MobileNetV2 by up to 10$\times$ and
MAdds (MAdds refers to Multiply-Adds as in
MobileNetV2~\cite{sandler2018mobilenetv2}) by about $25\%$, while
  maintaining the {\em same} accuracy. Additionally, due to its
  general formulation, \rpool can replace pooling layers anywhere in
  the architecture. For example, it can replace the final average pool
  layer in MobileNetV2 and improve accuracy by $\sim1\%$.

Finally, we modify the S3FD~\citep{zhang2017s3fd} architecture with
\rpool to construct an accurate face detection model which needs only
225 KB RAM -- small enough to be deployed on a Cortex-M4 based device
-- and achieves 0.78 MAP on the medium category of the WIDER FACE
dataset~\citep{yang2016wider} using 80$\times$ fewer MAdds than
EXTD~\citep{yoo2019extd} -- a state-of-the-art resource-constrained
face detection method.

In summary, we make the following contributions:
\begin{itemize}[leftmargin=*]\vspace{-1mm}
    \itemsep 0pt
    \topsep 0pt
    \parskip 2pt
\item A novel pooling operator that can rapidly down-sample input in a
  variety of standard CNN architectures, e.g., MobileNetV2,
  DenseNet121 while retaining the expressiveness.
\item Demonstrate that \rpool can reduce working memory and compute
  requirements for image classification and Visual Wake Words
  significantly while retaining comparable accuracy.
\item By combining \rpool with S3FD, we obtain a state-of-the-art face
  detection model for ARM Cortex-M4 class devices.
\end{itemize}

\section{Related Work}
\label{sec:rw}
\textbf{Pooling:} Max-pooling, average-pooling and strided convolution
layers~\citep{lecun2015deep} are standard techniques for feature
aggregation and for reducing spatial resolution in CNNs. Existing
literature on rethinking pooling~\citep{zhao2018multiactivation,
  he2015spatial, gong2014multi} focuses mainly on increasing accuracy
and does not take compute/memory efficiency into consideration which
is the primary focus of this paper.

\textbf{Efficient CNN architectures:} Most existing research on
efficient CNN architectures aims at reducing model size and number of
operations per inference. These methods include designing new
architectures such as DenseNet~\citep{huang2017densely},
MobileNets~\citep{howard2017mobilenets, sandler2018mobilenetv2} or
searching for them (ProxylessNAS~\citep{cai2018proxylessnas},
EfficientNets~\citep{tan2019efficientnet},
SqueezeNAS~\citep{shaw2019squeeze}). These architectures do not
primarily optimize for the peak working memory, which is a critical
constraint on devices powered by tiny microcontrollers. Previous work
on memory-optimized inference manipulates existing convolution
operator by reordering computations~\citep{cho2017mec, lai2018cmsis}
or performing them in place~\citep{gural2019memory}. However, most of
these methods provide relatively small memory savings and are
validated on low-resolution images like
CIFAR-10~\citep{krizhevsky2009learning}. Channel
pruning~\citep{he2017channel} is a method that tries to reduce memory
requirement by pruning out multiple convolution kernels in every
layer. While effective, channel/filter pruning does not tackle gradual
spatial downsampling and thus is a complementary technique to \rpool.

\textbf{Visual Wake Words:} Visual cues (visual wake word) to
``wake-up" AI-powered home assistant devices require real-time
inference on relatively small devices. \citet{chowdhery2019visual}
proposed a Visual Wake Words dataset and a resource-constrained
setting to evaluate various methods. Section~\ref{sec:rp+vww}
discusses the efficient \rpool based models and their performance for
this task.

\textbf{Face-detection on tiny devices:} Recent work including
EXTD~\citep{yoo2019extd}, LFFD~\citep{he2019lffd},
FaceBoxes~\citep{zhang2017faceboxes} and EagleEye~\citep{zhao2019real}
address the problem of accurate real-time face detection on
resource-constrained devices. EXTD and LFFD are the most accurate but
have high compute and memory requirements. On the other hand, EagleEye
and FaceBoxes have lower inference complexity but also suffer from
lower MAP scores. Face detection using \rpool is discussed in
Section~\ref{sec:fdexpts}.

\textbf{RNNs for Computer Vision}:
\label{sec:renet}
RNNs have been successful for sequential tasks but haven't been
extensively explored in the context of computer vision. An early work,
ReNet~\citep{visin2015renet}, uses RNN based layer as a replacement
for a convolution layer but does not aim at improving
efficiency. \rpool contrasts with ReNet as follows:
\vspace{-1mm}
\begin{enumerate}[leftmargin=*, label=\alph*.]
    \itemsep 0pt
    \topsep 0pt
    \parskip 0pt
    \item ReNet is designed to replace a convolutional layer by
      capturing the global context and leaves the local context to be
      captured by flattening non-overlapping patches. \rpool, on the
      other hand, uses overlapping patches and strongly captures local
      features and relies on subsequent standard convolutions to
      capture the global context. Hence, \rpool and ReNet are
      complementary methods and can be combined.
    \item Semantically, \rpool is a generalized pooling operator and
      can replace any pooling layer or strided convolution. However,
      ReNet does not correspond to any pooling abstraction, making it
      hard to combine with existing CNN models. For example, \rpool
      can modify S3FD architecture to achieve state-of-the-art
      real-time face detection with < 1 MB RAM while ReNet fails to
      fit in that context as a replacement layer since the receptive
      field of the output of ReNet layer varies across spatial
      positions.
    \item ReNet can still be used as a rapid downsampling
      layer. Table~\ref{tab:pooling} shows that \rpool outperforms
      ReNet with lower model size and fewer MAdds across datasets and
      architectures. E.g. ReNet+MobileNetV2 applied to ImageNet-1K is
      almost 4\% less accurate than \rpool+MobileNetV2, despite the
      same working RAM requirement and more MAdds per inference.
\end{enumerate}
\vspace{-1mm}
Inside-Outside Net~\citep{Bell_2016_CVPR} uses a ReNet based layer for
extracting context features in object detection while
PiCANet~\citep{Liu_2018_CVPR} uses it as a global attention function
for salient object detection. L-RNN~\citep{xie2016layer} inserts
multiple ReNet based layers but in a cascading fashion. See
Appendix~\ref{sec:renetapp} for more discussion.

PolygonRNN~\citep{acuna2018efficient}, CNN-RNN~\citep{wang2016cnn} and
Conv-LSTM~\citep{xingjian2015convolutional} also use RNNs in their
architectures but only to model certain sequences in the respective
tasks rather than tackling pooling and efficiency.

\section{What is \rpool?}
\label{sec:def}

Consider the output of an intermediate layer in a CNN of size $R\times
C \times f$, where $R$ and $C$ are the number of rows and columns and
$f$ is the number of channels. A typical $2\times 2$ pooling layer
(e.g. max or average) with stride $2$ would halve the number of rows
and columns. So, reducing dimensions by a factor of $4$ would
require {\em two} such blocks of convolutions and pooling. Our goal is to reduce the activation of
size $R\times C \times f$ to, say, $R/4 \times C/4 \times f'$ or
smaller in a single layer while retaining the information necessary
for the downstream task. We do so using an \rpoollayer illustrated in
Figure~\ref{fig:rnnpoolillus} that utilizes strided \rpool operators.

\subsection{The \rpool Operator and the \rpoollayer}

An \rpool operator of size $(r,c,k,h_1,h_2)$ takes as input an
activation patch of size $r\times c \times k$ corresponding to $k$
input channels, and uses a pair of RNNs -- $\mathrm{RNN}_1$ of hidden
dimension $h_1$ and $\mathrm{RNN}_2$ with hidden dimension $h_2$ -- to
sweep the patch horizontally and vertically to produce a summary of
size $1\times 1\times 4h_2$.

\begin{wrapfigure}{r}{0.5\columnwidth}
  \begin{minipage}{0.5\columnwidth}
  \vspace{-13pt}
\begin{algorithm}[H]
\caption{\rpool Operation}
\label{alg:rpool}
\small
\begin{algorithmic}[1]
\Require{$\mathbf{X}: [\mathbf{x}_{1,1} \dots \mathbf{x}_{r,c}]; \mathbf{x}_{i,j}\in \mathcal{R}^k$} 
\Ensure{$\rpool(\mathbf{X})$}
\Statex
\Function{$\mathrm{FastGRNN}$}{$\mathcal{P},\mathbf{x}$}
    \State{$[\mathbf{W},\mathbf{U},\mathbf{b}_z,\mathbf{b}_h] \gets \mathcal{P} $, $\mathbf{h}_0$ $\gets$ $\mathrm{randn}$} 
    \For{$k \gets 1$ to length$(\mathbf{x})$}
        \State{$\mathbf{z}$ $\gets$ $\sigma(\mathbf{Wx}_{k} + \mathbf{Uh}_{k-1} + \mathbf{b}_z)$}
        \State{$\tilde{\mathbf{h}}_{k}$ $\gets$ $\tanh(\mathbf{Wx}_{k} + \mathbf{Uh}_{k-1} + \mathbf{b}_h)$}
        \State{${\mathbf{h}}_{k}$ $\gets$ $\mathbf{z} \odot \mathbf{h}_{k-1} + (\mathbf{1}-\mathbf{z})\odot \tilde{\mathbf{h}}_{k}$}
    \EndFor
    \State \Return {$\mathbf{h}_T$}
\EndFunction
\Statex
\State{$\mathrm{RNN}_i(\_) \gets \mathrm{FastGRNN}(\mathcal{P}_i, \_)$, for $i\in \{1,2\}$}


\Function{\rpool}{$\mathbf{X}$}

        \State{$\mathbf{p}^{r}_{i}$ $\gets$ $\mathrm{RNN}_1(\mathbf{X}_{i , 1\le j \le c})$, for all $1\leq i\leq r$} \label{lst:line:1}
      
   
    \State{$\mathbf{q}^{r_1}$ $\gets$ $\mathrm{RNN}_2(\mathbf{p}^{r}_{1\le i \le r})$} \label{lst:line:2}
    
    \State{$\tilde{\mathbf{p}}^{r}$ $\gets$ $\mathrm{reverse}({\mathbf{p}}^{r})$, $\mathbf{q}^{r_2}$ $\gets$ $\mathrm{RNN}_2( \tilde{\mathbf{p}}^{r}_{1\le i \le r})$}  \label{lst:line:3}
    
    \Statex
      
        \State{$\mathbf{p}^{c}_{j}$ $\gets$ $\mathrm{RNN}_1(\mathbf{X}_{1\le i \le r, j})$, for all  $1\leq j\leq c$}\label{lst:line:4}
      
    
    \State{$\mathbf{q}^{c_1}$ $\gets$ $\mathrm{RNN}_2(\mathbf{p}^{c}_{1\le j \le c})$}\label{lst:line:5}
        
    \State{$\tilde{\mathbf{p}}^{c}$ $\gets$ $\mathrm{reverse}(\mathbf{p}^{c})$, $\ \mathbf{q}^{c_2}$ $\gets$ $\mathrm{RNN}_2(\mathbf{\tilde{p}}^{c}_{1\le j \le c})$}
        \label{lst:line:6}
      
    \Statex

    \State \Return {$[\mathbf{q}^{r_1}, \mathbf{q}^{r_2}, \mathbf{q}^{c_1}, \mathbf{q}^{c_2}]$}
\EndFunction
\end{algorithmic}
\end{algorithm}
\vspace{-35pt}
\end{minipage}
\end{wrapfigure}

Algorithm \ref{alg:rpool} describes the
\rpool operator wich applies two parallel pipelines to a patch and
concatenates their outputs. In the first, $\mathrm{RNN}_1$ traverses
each row and summarizes the patch horizontally (Line~\ref{lst:line:1})
and then $\mathrm{RNN}_2$ trverses the outputs of $\mathrm{RNN}_1$
(Lines~\ref{lst:line:2}-\ref{lst:line:3}) bi-directionally. In the
second pipeline $\mathrm{RNN}_1$ first traverses along columns to
summarize the patch vertically (Line~\ref{lst:line:4}) and then
$\mathrm{RNN}_2$ (Lines~\ref{lst:line:5}-\ref{lst:line:6}) summarizes
bi-directionally.

While it is possible to use GRU~\citep{cho2014learning} or
LSTM~\citep{hochreiter1997long} for the two instances of RNN in \rpool,
we use FastGRNN~\citep{kusupati2018fastgrnn} for its compact size and
fewer MAdds (see Appendix~\ref{sec:ablation}).

An \rpoollayer consists of a single \rpool operator strided over an
input activation map and takes as input two more parameters: patch
size and the stride length.  Note that there are only two RNNs
($\mathrm{RNN}_1$ \& $\mathrm{RNN}_2$) in an \rpool operator, thus
weights are shared for both the row-wise and column-wise
passes~($\mathrm{RNN}_1$) and all bi-directional
passes~($\mathrm{RNN}_2$) across every instance of \rpool in an
\rpoollayer.

\subsection{Probing the Efficacy of \rpool}

\textbf{Capturing edges, orientations, and shapes:} To demonstrate the
capabilities of RNNs as spatial operators for vision tasks such as
capturing edges, orientations, and shapes, we performed experiments on
synthetic data. \rpool learns how to capture edges, orientations, and
shapes as effectively as convolutional layers which reinforces the
choice of RNNs as spatial operators. Appendix~\ref{sec:edges} provides
further details of these experiments.

\textbf{Comparing performance with pooling operators:} We also
performed experiments to contrast the down-sampling power of \rpool
against standard pooling operators on
CIFAR-10~\citep{krizhevsky2009learning}. As discussed in Appendix~\ref{sec:cifar}, \rpool significantly outperforms standard pooling operators in terms of accuracy.

\vspace{-5pt}
\section{How to use the \rpoollayer?}
\label{sec:usage}
\vspace{-5pt}

\begin{wrapfigure}{r}{0.48\columnwidth}
  \centering
  \vspace{-25pt}
    \includegraphics[width=0.48\columnwidth]{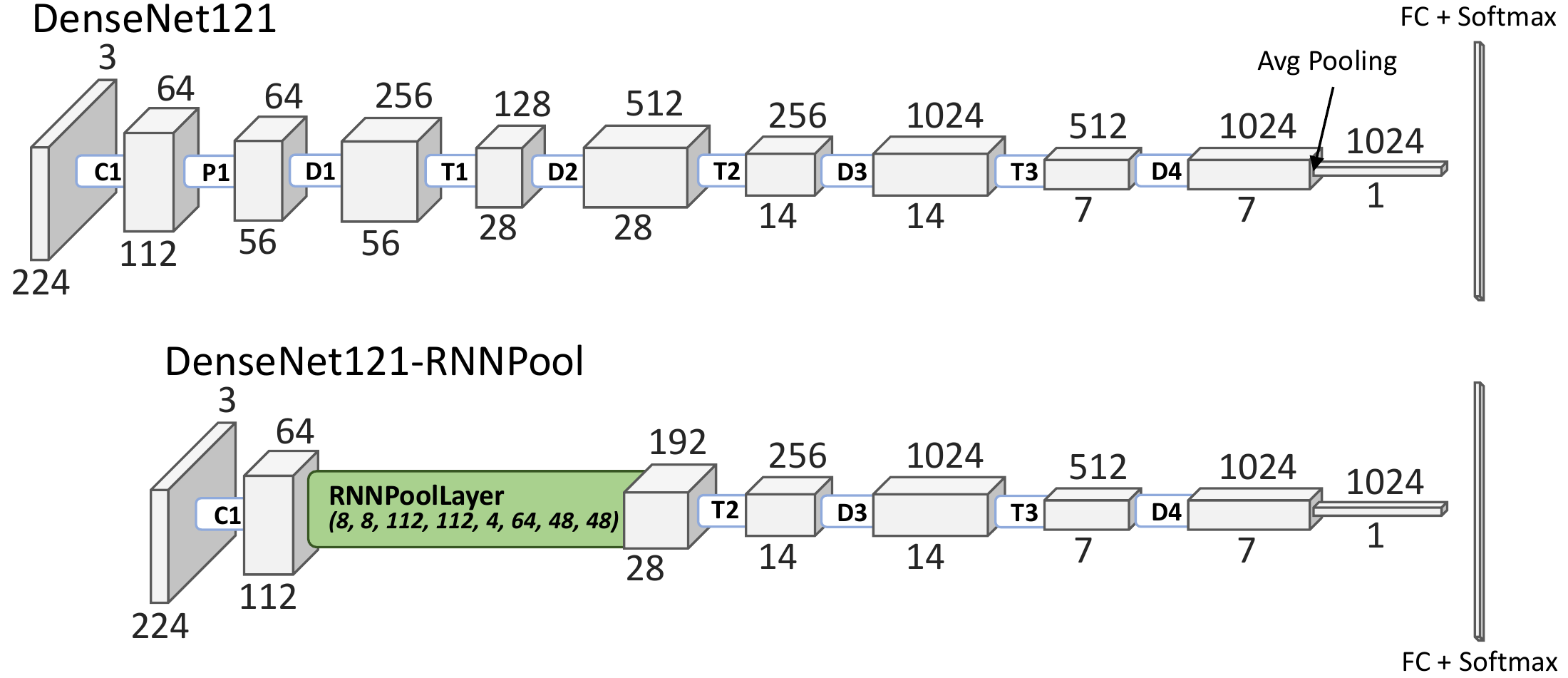}
\caption{\small \textbf{DenseNet121-\alg}: obtained by replacing P1,
  D1, T1 and D2 blocks in DenseNet121 with an \rpoollayer.}
\label{fig:DenseNet-PRNN}
\vspace{-12pt}
\end{wrapfigure}

\rpool can be used to modify standard CNN architectures and reduce
their working memory as well as computational requirements. Typically,
such modifications involve replacing one or more stacks of
convolutions and pooling layers of the ``base'' (original)
architecture with an \rpoollayer and retraining from scratch. We
describe architecture modification strategies here and demonstrate
their effectiveness through extensive experimentation in
Section~\ref{sec:expts}.

\textbf{Replacement for a Sequence of Blocks:} Consider the
DenseNet121~\citep{huang2017densely} architecture in
Figure~\ref{fig:DenseNet-PRNN}.  It consists of one convolutional
layer, followed by repetitions of ``Dense'' (D), ``Transition'' (T)
and ``Pooling'' (P) blocks which gradually reduce the size of the
image while increasing the number of channels.  Of all these layers,
the first block following the initial convolutional layer (D1)
requires the most working memory and compute as it works on large
activation maps that are yet to be down-sampled.  Further, the
presence of 6 layers within each dense block makes it harder to work
with small memory (see Proposition~\ref{prop:mem}).  This is also true
of other architectures such as MobileNetV2, EfficientNet, and ResNet.

We can use an \rpoollayer to rapidly down-sample the image size and
bypass intermediate large spatial resolution activations. In
DenseNet121, we can replace 4 blocks - P1, D1, T1, D2 - spanning
\emph{39} layers with a single \rpoollayer to reduce the activation
map from size $112\times 112\times64$ to $28\times28\times128$ (see
Figure~\ref{fig:DenseNet-PRNN}). The replacement \rpoollayer can be
executed patch-by-patch without re-computation, thus reducing the need
to store the entire activation map across the image.  These two
factors greatly reduce the working memory size as well as the number
of computations. DenseNet121-\rpool achieves an accuracy of $94.8\% $
on ImageNet-10 (see Appendix~\ref{sec:datasets} for dataset details)
which is comparable to $95.4\%$ of the original DenseNet121 model.

A similar replacement of functional blocks with \rpoollayer can be
performed for MobileNetV2 as specified in
Table~\ref{tab:mobilenetv2-rp} of Appendix~\ref{sec:architectures},
and leads to a similar reduction in the size of the largest activation
map while retaining accuracy.  These results extend to other networks
like EfficientNet, ResNet and GoogLeNet~\citep{szegedy2015going},
where residual connection based functional blocks in the initial parts
can be effectively replaced with the \rpoollayer with improvements in
working memory and compute, while retaining comparable accuracy. These
results are listed in
Table~\ref{tab:10class}. Appendix~\ref{sec:ablation} presents further
ablation studies on \rpool and its base model.

\textbf{Replacement for Pooling Layers:} \rpool has the same input and
output interface as any pooling operator and hence, \rpoollayer can
replace any standard pooling layer while providing more accurate
aggregation.  For example, DenseNet121-\rpool has three pooling layers
one each in T2, T3, and the final average pool layer.
Table~\ref{tab:10class} shows that, on ImageNet-10,
DensetNet121-\rpool loses 0.6\% accuracy compared to its base
model. But, replacing all three remaining pooling layers in
DenseNet121-\rpool with a \rpoollayer results in almost the same accuracy
as the {\em base} DenseNet121 but with about 2$\times$ and 4$\times$
lower compute and RAM requirement respectively.  We can further drop
14 dense layers in D3 and 10 layers in D4 to bring down MAdds and RAM
requirement to 0.79G MAdds and 0.43 MB, respectively, while still
ensuring $94.2\%$ accuracy.

\label{sec:introface}
\textbf{Replacement in Face Detection models:} As in the above
architectures, we can use \rpoollayer to rapidly down-sample the image
by a factor of $4\times 4$ in the early phase of an S3FD face
detector~\citep{zhang2017s3fd}. The resulting set of architectures
(with different parameters) are described in Appendix
~\ref{sec:fdmodels}. For example, the \rpool-Face-Quant model has a
state-of-the-art MAP for methods that are constrained to at most 256
KB of working RAM (Table~\ref{tab:facedetection}).

\begin{table*}[t]
  \centering
  \caption{\small Comparison of inference complexity and accuracy with
    and without \rpoollayer on ImageNet-10.}
  
  \resizebox{\columnwidth}{!}{%
    \begin{tabular}{ l | c  c | c  c | c  c | c  c  c  c }
      \toprule
      \multicolumn{1}{l|}{\multirow{3}*{Model}} &  \multicolumn{6}{|c|}{Base} & \multicolumn{4}{|c}{\multirow{2}*{\alg}} \\\cmidrule{2-7}
      &  \multirow{2}*{\begin{tabular}[c]{@{}c@{}}Accuracy \\ (\%)\end{tabular}} & \multirow{2}*{Parameters} & \multicolumn{2}{|c|}{Memory Optimised} &  \multicolumn{2}{|c|}{Standard Calculation \cite{chowdhery2019visual,sandler2018mobilenetv2}}  \\\cmidrule{4-11}
      &  &  & Peak RAM  & MAdds & Peak RAM  & MAdds & Accuracy (\%) & Parameters & Peak RAM  & MAdds \\
      \midrule
      MobileNetV2 & 94.20 & ~2.20M & 0.38 MB & ~1.00G & 2.29 MB & 0.30G & \textbf{94.40} & \textbf{~2.00M} & \textbf{0.24 MB} & \textbf{0.23G} \\
      EfficientNet-B0 & 96.00 & ~4.03M & 0.40 MB & ~1.09G & 2.29 MB & 0.39G & \textbf{96.40} & \textbf{~3.90M} & \textbf{0.25 MB} & \textbf{0.33G}\\
      ResNet18 & \textbf{94.80} & 11.20M & 0.38 MB & 21.58G & 3.06 MB & 1.80G & 94.40 & \textbf{10.60M} & \textbf{0.38 MB} & \textbf{0.95G} \\
      DenseNet121 & \textbf{95.40} & ~6.96M & 1.53 MB & 24.41G & 3.06 MB & 2.83G & 94.80 & \textbf{~5.60M} & \textbf{0.77 MB} & \textbf{1.04G} \\
      GoogLeNet & \textbf{96.00} & ~9.96M & 1.63 MB & ~3.32G & 3.06 MB & 1.57G & 95.60 & \textbf{~9.35M} & \textbf{0.78 MB} &  \textbf{0.81G} \\
      \bottomrule
   \end{tabular}}
\vspace{-5mm}
  \label{tab:10class}
\end{table*}

\textbf{Inference memory requirements}: Computing exact memory and
compute requirement of a large CNN model is challenging as the
execution order of activations in various layers can be re-organized
to trade-off memory and compute. For example, in the {\em
  memory-optimized} column of Table~\ref{tab:10class} we present the
compute usage of a variety of baseline architectures when their
execution order (EO) is restricted to using no more memory than the
corresponding \rpool based architecture. That is, we identify the
memory bottleneck layers in various architectures whose activation map
size is almost same as that of the corresponding \rpool-based
model. We then compute every voxel of this layer by re-computing the
required set of convolutions, {\em without} storing them. CNNs, in
general, have significant compute requirement and such re-compute
intensive optimizations make the architecture infeasible even for
large devices, e.g. DenseNet121 requires 24.41G MAdds in this scheme
(Table~\ref{tab:10class}).

A standard approach is to restrict execution orders that do not
require any re-computation of intermediate activation maps. A
straightforward and standard EO is the one where the computation is
executed {\em
  layer-by-layer}~\citep{chowdhery2019visual,sandler2018mobilenetv2}.
The memory requirement of such a scheme would correspond to the
largest activation map in the architecture, except the
output of 1x1 convolution layers which can be computed on the
fly. This approach mimics the memory requirement of existing platforms
like TF-lite~\citep{tf-lite} and is proposed as a standard benchmark
for comparing resource-constrained inference
methods~\citep{chowdhery2019visual}.  Following this prior convention,
we list the inference complexity for various architectures under the
{\em compute-optimized} columns in Table~\ref{tab:10class}, unless the
operation is easy to compute on the fly like 1x1 convolution or
patch-by-patch computation of \rpool.  Appendix~\ref{sec:compute-opt}
provides more details about these calculations.

The above scheme is easy to implement and allows an inference pipeline
that is more modular and easy to debug and could allow faster
inference on neural network accelerators~\citep{tpu}. But, in
principle, one can design execution orders (EO) that do not re-compute
any intermediate layers, but are still not required to store entire
activation maps, especially the largest ones. So, a rigorous
quantification of the memory requirement of a model (without any
re-compute) needs to show that any valid execution order requires a
certain amount of working memory at some point in its execution, and
also demonstrate a valid EO with the same memory requirement as a
matching upper bound. We achieve this with the following proposition,
whose proof and corollaries are in Appendix~\ref{sec:space}.
\vspace{-1mm}
\begin{proposition}
	\label{prop:mem}
	Consider an $l$-layer ($l>1$) convolutional network with a
        final layer of size $m\times n$.  Suppose the for each node in
        the output layer, the size of receptive field in intermediate
        layer $q \in[l-1]$ is $(2k_q+1)\times(2k_q+1), k_q>0$ and that
        this layer has $c_q$ channels and stride $1$.  Any serial
        execution order of this network that disallows re-computation
        requires at least $2 \sum_{q=1}^{l-1} c_q k_q \times
        min(m-1,n-1)$ memory for nodes in the intermediate layers.
\end{proposition}
\vspace{-1mm}
The above proposition shows that for a CNN with receptive field $k_q$
at the $q$-th layer, the memory requirement scales linearly with the
height/width of the activation map and with the number of layers. As
networks like MobileNetV2 or DenseNet have blocks with a significant
number of convolution layers and large receptive field, this
proposition implies that it is not possible to significantly reduce
the memory requirement over the standard {\em layer-by-layer}
approach. For example, our un-optimized calculations for \rpool
architectures still give us $3-4$x reduction in peak RAM usage when
compared to the {\em minimum} RAM requirement of the corresponding
base architecture (see Appendix~\ref{sec:opt-mem}). Further, similar
optimization can be applied to \rpool based architectures, so the relative reduction in memory by \rpool does not change
significantly. The implications of the above proposition, i.e., the
peak memory of various networks without re-compute is calculated in
Appendix~\ref{sec:opt-mem}.

\begin{table*}[t]
  \centering
  \caption{\small Impact of various downsampling and pooling operators
    on the accuracy, inference complexity and the model size of three
    {\em base} architectures: MobileNetV2 and DenseNet121 for
    ImageNet-10 dataset, and MobileNetV2-0.35x for Visual Wake Word
    dataset. First block of the table represents the base network and
    a modified network where the last average pooling layer in the
    network is replaced by \rpoollayer. Second block represent
    modified networks where the image is passed through a convolution
    layer followed by various downsampling methods to reduce the size
    of image by a factor of $4\times4$. The last row represents the
    architecture from the second block with \rpoollayer with an
    additional \rpool replacing the last layer. Peak RAM usage
    computed using standard convention of~\citep{chowdhery2019visual}
    is the same for all methods in the second block. Note that
    \rpoollayer+Last layer \rpool has accuracy similar to the base
    network while other methods like ReNet are 2-3\% less accurate.}

    \vspace{-5pt}
    \resizebox{\columnwidth}{!}{%
    \begin{tabular}{ l | c  c  c | c  c  c | c  c  c }
      \toprule
      \multicolumn{1}{l|}{\multirow{3}*{{Method}}} & \multicolumn{6}{|c|}{\textbf{ImageNet-10}} &
      \multicolumn{3}{|c}{\textbf{Visual Wake Words}}\\\cmidrule{2-10}
      & \multicolumn{3}{|c|}{MobileNetV2} & \multicolumn{3}{|c|}{DenseNet121}
      & \multicolumn{3}{|c}{MobileNetV2-0.35$\times$} \\\cmidrule{2-10}
      & Accuracy (\%) & MAdds & Parameters & Accuracy (\%) & MAdds & Parameters & Accuracy (\%) & MAdds & Parameters\\
      \midrule
      Base Network & {94.20} & 0.300G & 2.2M & \textbf{95.40}& 2.83G & 6.96M & {90.20} & 53.2M & 296K\\
      Last layer \rpool & 95.00 & 0.334G & 2.9M & \textbf{95.40} & 3.05G & 7.41M & \textbf{91.14} & 53.4M & 300K\\\midrule
      Average Pooling & 90.80 & \textbf{0.200G} & \textbf{2.0M} & 92.80 & \textbf{0.71G} & \textbf{5.59M} & 86.85 & \textbf{31.9M} & \textbf{255K}\\
      Max Pooling & 92.80 & \textbf{0.200G} & \textbf{2.0M} & 93.40 & \textbf{0.71G} & \textbf{5.59M} & 86.92 & \textbf{31.9M} & \textbf{255K}\\
      Strided Convolution & 93.00 & 0.258G & 2.1M & 93.80 & 1.33G & 6.38M & 88.08 & 39.2M & 264K\\
      ReNet & 92.20 & {0.296G} & {2.3M} & 93.00 & {1.35G} & {6.41M} & 88.10 & {46.4M} & {277K}\\
      \rpoollayer & {94.40} & 0.226G & \textbf{2.0M} & {94.80} & 1.04G & 5.60M & { 89.57} & 37.7M & \textbf{255K}\\\midrule
      \begin{tabular}[c]{@{}l@{}}\rpoollayer +\\ Last layer \rpool\end{tabular} & \textbf{95.60} & 0.260G & 2.7M & 95.00 & 1.26G & 6.06M & 89.65 & 37.9M & 259K\\
      \bottomrule
  \end{tabular}}
  \label{tab:pooling}
\end{table*}

\section{Evaluation of \alg on Vision Tasks}
\label{sec:expts}

We present empirical evidence that \rpool operator is compatible with
popular CNN architectures for vision tasks, and can push the envelope
of compute/memory usage vs accuracy curve. Further, we show
that \rpool combined with MobileNetV2~\citep{sandler2018mobilenetv2}
generates accurate models for Visual wake words and face detection
problems that can be deployed on tiny Cortex-M4 microcontrollers. See
Appendix~\ref{sec:hyperparams} for more details about model training
and hyperparameters used for the experiments.

\subsection{\alg for Image Classification}\label{sec:exp_class}

We first focus on ImageNet-10, a 10 class subset of
ImageNet-1K~\citep{deng2009imagenet} where the classes correspond to
the categories in CIFAR-10~\citep{krizhevsky2009learning}.  We study
this dataset because in several realistic tiny devices scenario, like
intrusion detection, we are interested in identifying the presence/absence
of a few, rather than 1000, classes of objects. The dataset is divided
into 1300 images for training and 50 for validation per class.  More
details and rationale about the dataset can be found in the
Appendix~\ref{sec:datasets}.

Table~\ref{tab:pooling} compares \rpoollayer against other standard
pooling operators as used in MobileNetV2 and DenseNet121 base networks
(see Appendix~\ref{sec:arch-classification} for description of the
architecture).  It shows that with the same memory usage, \rpool is up
to $4$\% more accurate than the standard pooling operators. While
standard pooling operators are cheaper than \rpool, the overall
compute requirement of \rpool based architectures is similar to
pooling based architectures. Furthermore, replacing the last average
pooling layer in the base network with \rpool further increases
accuracy, thus demonstrating the flexibility
of \rpoollayer. Table~\ref{tab:pooling} also contrasts \rpool with
ReNet~\citep{visin2015renet} as a downsampling layer. We observe
that \rpool is a much better alternative for downsampling layers in
terms of accuracy (better by up to 2\%), model size, and MAdds for the
same amount of working memory.

Next, we study the compatibility of \rpool with different
architectures. Table~\ref{tab:10class} shows that \rpool based
architectures maintain the accuracy of base models while significantly
decreasing memory and compute requirement. See Section \ref{sec:usage}
and Appendix \ref{sec:peak} for a discussion on the calculation of
memory and compute requirements of different models.

\begin{wraptable}{l}{0.5\columnwidth}
\vspace{-3mm}
  \centering
    \caption{\small Comparison of resources and accuracy with MobileNets for ImageNet-1K.}
  \resizebox{0.5\columnwidth}{!}{%
    \begin{tabular}{ l  c  c  c  c }
      \toprule
      \multicolumn{1}{l}{Method} & {Peak RAM} & {Parameters} & {MAdds} & {Accuracy (\%)} \\
      \midrule
      MobileNetV1 & 3.06MB & 4.2M & 569M & \textbf{69.52}\\
      MobileNetV1-ReNet & 0.77MB & 4.2M & 487M & 66.90\\
      MobileNetV1-\alg & \textbf{0.77MB} & \textbf{4.1M} & \textbf{417M} & 69.39\\
      \midrule
      MobileNetV2 & 2.29MB & 3.4M & 300M & \textbf{71.81}\\
      MobileNetV2-ReNet & 0.24MB & 3.6M & 296M & 66.72\\
      MobileNetV2-\alg & \textbf{0.24MB} & \textbf{3.2M} & \textbf{226M} & 70.14\\
      \bottomrule
  \end{tabular}}
  \label{tab:imagenet}
\vspace{-3mm}
\end{wraptable}

Finally, Table~\ref{tab:imagenet} presents results on the complete
ImageNet-1K~\citep{deng2009imagenet} dataset with MobileNetV1 and
MobileNetV2 as the base architectures. ReNet and \rpool based models
are constructed in a manner similar to the models in
Table~\ref{tab:10class}. See Table~\ref{tab:mobilenetv2-rp} for the
complete specification of the MobileNetV2+\rpool model.
MobileNetV1+\rpool model is constructed similarly with $h_1=h_2=16$.
Consistent with the results on ImageNet-10, \rpool retains almost same
accuracy as the base models while decreasing memory usage
significantly. Furthermore, \rpool based models are also $3-4$\% more
accurate than {\em ReNet based models}. In this work, we focus on
state-of-the-art resource-constrained models that do not require
neural architecture search (NAS); we leave extension of \rpool for NAS
based architectures like EfficientNets~\citep{tan2019efficientnet} for
future work.

\subsection{\alg for Visual Wake Words}
\label{sec:rp+vww}
\begin{wrapfigure}{r}{0.6\columnwidth}
\vspace{-8mm}
 	\centering
  \includegraphics[width=0.6\columnwidth]{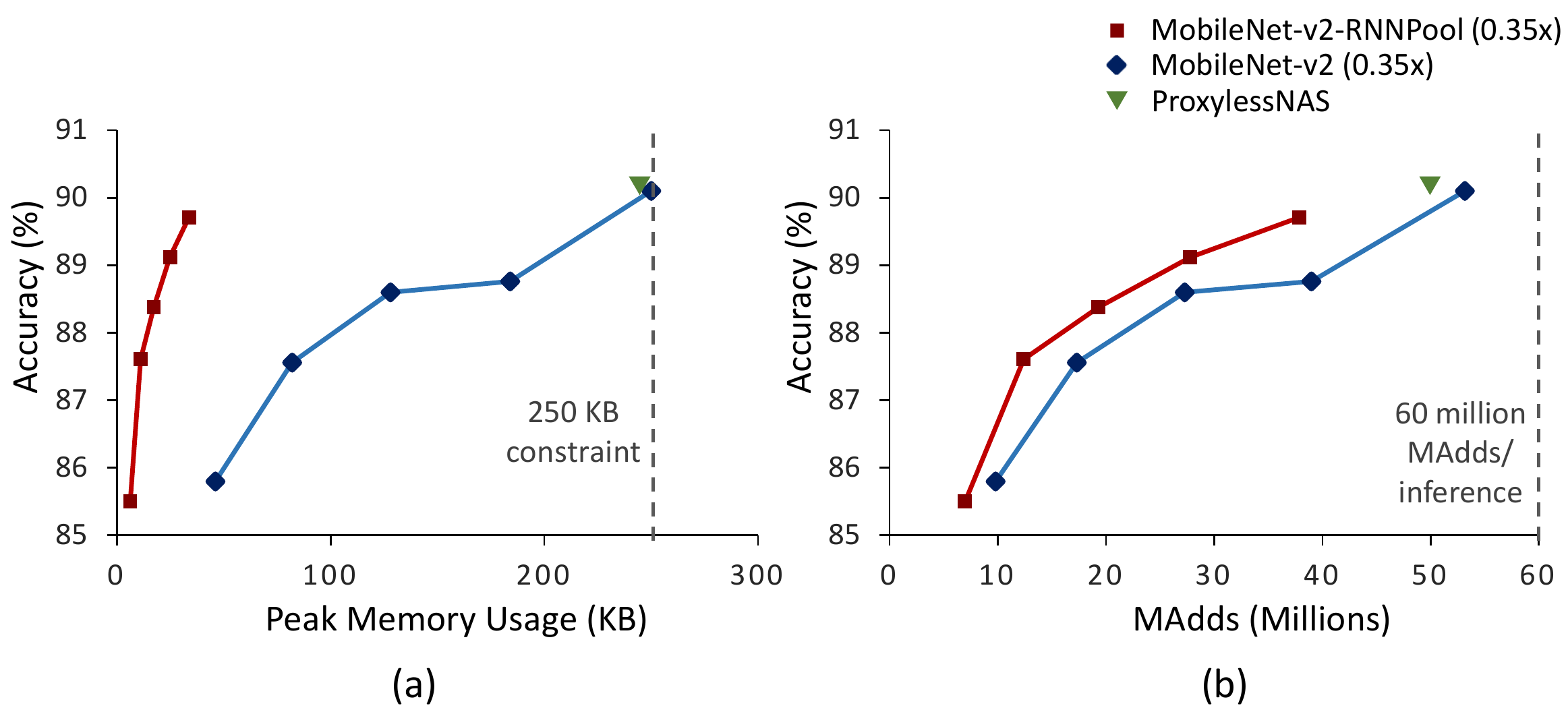}
  \caption{\small Visual Wake Word: MobileNetV2-\rpool requires
    $8\times$ less RAM and 40\% less compute than baselines. We cap
    the number of parameters at $\leq 250$K instead of the $290$K
    allowed by MobileNetV2 (0.35$\times$). ProxylessNAS has $242$K
    parameters.}
  \label{fig:vww}
\vspace{-4mm}
\end{wrapfigure}

The Visual Wake Words challenge~\citep{chowdhery2019visual} presents a
relevant use case for computer vision on tiny microcontrollers. It
requires detecting the presence of a human in the frame with very
little resources --- no more than 250 KB peak RAM usage and model
size, and no more than 60M MAdds/image. The existing state-of-the-art
method~\citep{chowdhery2019visual} is MobileNetV2-$0.35\times$ with 8
channels for the first convolution and 320 channels for the last
convolution layer.  We use this as our baseline and replace
convolutions with an \rpoollayer. After training a floating-point
model with the best validation accuracy, we perform per-channel
quantization to obtain 8-bit integer weights and activations.

Table~\ref{tab:pooling} compares the accuracy of the baseline and new
architectures on this task. Replacing the last average pool layer
with \rpool increases the accuracy by $\ge$ 1\%.  Inserting \rpool
both at the beginning of the network and at the end provides a model
whose accuracy is within $0.6$\% of the baseline but with far smaller
memory requirement (250 $\to$ 33.68 KB), model size, and MAdds.  Peak
memory usage is calculated using the same convention
as~\citep{chowdhery2019visual}.

Further, we sweep across input image resolutions of \{96, 128, 160,
192, 224\} to trade-off between accuracy and
efficiency. Figure~\ref{fig:vww} shows that \rpool models are
significantly cheaper during inference in terms of compute and memory
while offering the same accuracy as the baselines. For example, peak
memory usage of MobileNetV2-0.35$\times$ with the lowest resolution
images is $\sim$40 KB, while our model requires only 34 KB RAM despite
using the highest resolution image and providing $\sim$4\% higher
accuracy. Note that ProxylessNAS~\citep{mit-hanvww} was the winner of
the Visual Wake Words challenge. We report it's accuracy on the final
network provided by the authors. To be consistent, we train the model
only on the training data provided, instead of pretraining with
ImageNet-1K used by ProxylessNAS in the wake word challenge.

\subsection{\alg for Face Detection}
\label{sec:fdexpts}

\begin{table*}[t]
  \centering
  \caption{\small Comparison of memory requirement, no. of
    parameters and validation MAP of various Face Detection
    architectures when applied to $640\times 480$ RGB images from the Wider Face dataset. \alg-Face-C achieves higher accuracy
    than the baselines despite using 3$\times$ less RAM and
    4.5$\times$ less MAdds. \alg-Face-Quant enables deployment on
    Cortex-M4 class devices with 6-7\% accuracy gains over the
    cheapest baselines.}

\resizebox{\columnwidth}{!}{%
	\begin{tabular}{ l | c  c  c | c  c  c | c  c  c }
		\toprule
		\multicolumn{1}{l|}{\multirow{2}*{Method}} & \multirow{2}*{Peak RAM} & \multirow{2}*{Parameters} & \multirow{2}*{MAdds} & \multicolumn{3}{|c|}{MAP}& \multicolumn{3}{|c}{MAP  for $\leq$ 3 faces}\\\cmidrule{5-10}
		&  &  &  & Easy & Medium & Hard & Easy & Medium & Hard\\
		\midrule
		EXTD & 18.75 MB & \textbf{0.07M} & 8.49G & 0.90 & 0.88 & \textbf{0.82} & 0.93 & 0.93 & 0.91\\
		LFFD & 18.75 MB & 2.15M & 9.25G & 0.91 & 0.88 & 0.77 & 0.83 & 0.83 & 0.82\\
		\alg-\ensuremath{{\rm Face}}-\ensuremath{{\rm C}} & \textbf{~6.44 MB} & 1.52M & \textbf{1.80G} & \textbf{0.92} & \textbf{0.89} & 0.70 & \textbf{0.95} & \textbf{0.94} & \textbf{0.92}\\
		\midrule
		FaceBoxes & 1.76 MB & \textbf{1.01M} & 2.84G & 0.84 & 0.77 & 0.39 & - & - & -\\
		\alg-\ensuremath{{\rm Face}}-\ensuremath{{\rm B}} & 1.76 MB & 1.12M & \textbf{1.18G} & \textbf{0.87} & \textbf{0.84} & \textbf{0.67} & 0.91 & 0.90 & 0.88\\
		\midrule
        EagleEye & 1.17 MB & 0.23M & \textbf{0.08G} & 0.74 & 0.70 & 0.44 & 0.79 & 0.78 & 0.75\\
		\alg-\ensuremath{{\rm Face}}-\ensuremath{{\rm A}} & 1.17 MB & \textbf{0.06M} & 0.10G & {0.77} & {0.75} & {0.53} & {0.81} & {0.79} & {0.77}\\
		\alg-\ensuremath{{\rm Face}}-\ensuremath{{\rm Quant}} & \textbf{225 KB} & 0.07M & 0.12G & \textbf{0.80} & \textbf{0.78} & \textbf{0.53} & \textbf{0.84} & \textbf{0.83} & \textbf{0.81}\\
		\bottomrule
	\end{tabular}}
	\label{tab:facedetection}
\end{table*}

We experiment with multiple architectures we call \rpool-Face-* for
face detection suggested in Section~\ref{sec:introface} and described
in greater detail in Appendix~\ref{sec:fdmodels}.  We train and
validate these architectures with the WIDER FACE
dataset~\citep{yang2016wider}. Versions Quant, A, B, and C of the \rpool-Face
use \rpoollayer of hidden dimensions 4, 4, 6 and 16,  
respectively.

Table~\ref{tab:facedetection} compares validation Mean Average
Precision (MAP) for easy, medium, and hard subsets. MAP is a standard
metric for face detection and measures the mean area under the
precision-recall curve.
We report MAP scores for baselines
based on the official open-source code or pre-trained models. For
Eagle-Eye~\citep{zhao2019real}, we re-implemented the method as the
source code was not available. For EXTD~\citep{yoo2019extd}, we report
MAdds of the EXTD-32 version - the computationally cheapest. EXTD and
LFFD~\citep{he2019lffd} are accurate but are computationally
expensive. In contrast, \rpool-Face-C achieves better MAP in the easy
and medium subsets despite using $\sim4.5\times$ less compute and
$\sim3\times$ less RAM.

FaceBoxes~\citep{zhang2017faceboxes} and Eagle-Eye reduce MAdds and
peak memory usage by aggressively down-sampling the image or by decreasing
the number of channels leading to inaccurate models. In
contrast, \rpool-Face-A and \rpool-Face-B achieve significantly higher
MAPs than these methods while still ensuring smaller MAdds and peak
RAM usage. We also compare MAP scores for images that have $\leq$ 3
faces, which is a more realistic face-detection setting for
tiny devices. Here also, \rpool-Face-C is more
accurate than all the baselines. Finally, \rpool-Face-Quant uses byte quantization to
reduce the model size so it can be deployed on Cortex-M4 devices which
typically have $\leq 256$ KB RAM, while still having $>0.80$ MAP
accuracy on images with $\leq 3$ faces. See
Appendix~\ref{sec:facedetviz} for a qualitative evaluation of our
method against the baselines.

\subsection{\rpool based Model for ARM Cortex-M4 Microcontrollers}
Finally, we develop a face detection model for conference/class room settings that can be deployed on ARM Cortex-M4 class devices. To this end,  we develop a
more compact version of the face detection model, \rpool-Face-M4
(Table \ref{tab:rpool-m4} in Appendix~\ref{sec:fdmodels}), which has
only 4 MBConv blocks. For further reduction in MAdds and model-size, we train the \rpool parameters to be sparse. That is, 
$\mathbf{W}$ matrix of $\mathrm{RNN}_1$ is 50\% non-zeros while the rest
of the matrices in \rpool are 30\% non-zeros.

To not overshoot RAM for storing input image, we use
320$\times$240$\times$1 monochrome images for training and testing.
For evaluation, we first train on the WIDER FACE dataset and then
fine-tune on the SCUT-HEAD dataset~\citep{peng2018detecting} which consists of images in conference/class rooms. We then
use the SeeDot~\citep{gopinath2019compiling} compiler to quantize our
model to 8 bits and generate C code for
deployment. Table~\ref{tab:m4vsmobssd} compares the resource
requirements and MAP on the SCUT-HEAD validation set (random 80\%-20\% split)
of \rpool-Face-M4 against a similarly trained MobileNetV2-SSDLite
model which is a state-of-the-art architecture for low-cost detection.

\begin{table}[t]
\centering
    \caption{\small Comparison of resources and MAP on the SCUT-HEAD dataset. \rpool-Face-M4 can be effectively deployed on an M4 device with $<$256 KB RAM in contrast to MobileNetV2-SSDLite low-cost detection model.}
\resizebox{0.7\columnwidth}{!}{%
\begin{tabular}{l c c c c c}
  \toprule
  Model  & MAP &  Peak RAM &  MAdds & Model Size\\
  \midrule
  MobileNetV2-SSDLite & \textbf{0.63} & 3.51 MB & 540M & 11.32 MB\\ 
  \rpool-Face-M4 & 0.58 & \textbf{188 KB} & \textbf{70M} & \textbf{160 KB}\\ 
  \bottomrule
\end{tabular}}
\label{tab:m4vsmobssd}
\vspace{-2mm}
\end{table}

Note that MobileNetV2-SSDLite cannot be deployed on a Cortex-M4 device
even with 8-bit quantization as the peak RAM requirement is much more
than the 256 KB limit of the device. \rpool-Face-M4 model processes a
single image in 10.45 seconds on an ARM Cortex-M4 microcontroller
based STM32F439-M4 device clocked at 168 MHz.

\section{Conclusions}
\label{sec:conc}
In this paper, we proposed \rpool, an efficient RNN-based pooling
operator that can be used to rapidly downsample activation map sizes
thus significantly reduce inference-time memory and compute
requirements for a variety of standard CNNs. Due to syntax level
similarity with pooling layers, we can use \rpool in most existing CNN
based architectures.  These replacements retain accuracy for tasks
like image classification and visual wake words. Our S3FD based \rpool
model for face detection provided accurate models that can be deployed
on tiny Cortex-M4 microcontrollers. Finally, we showed with
Proposition~\ref{prop:mem} that calculations of minimum memory
requirement for standard architectures can be made rigorous and
demonstrate that despite such optimizations of standard CNNs, \rpool
based models can be significantly more efficient in terms of
inference-time working memory. Using neural architecture search for
\rpool based models to further reduce inference cost is an immediate
and interesting direction.

\section*{Broader Impact}
\label{sec:broad}

\textbf{Pros:} ML models are compute-intensive and are
typically served on power-intensive cloud hardware with a large
resource footprint that adds to the global energy footprint.  Our models
can help reduce this footprint by (a) allowing low power edge sensors
with small memory to analyze images and admit only interesting images
for cloud inference, and (b) reducing the inference complexity of
the cloud models themselves.  Further, edge-first inference enabled by our
work can reduce reliance on networks and also help provide privacy
guarantees to end-user.  Furthermore, vision models on tiny edge
devices enables accessible technologies, e.g., Seeing AI~\citep{seeingai} for
people with visual impairment.

\textbf{Cons:} While our intentions are to enable socially valuable
use cases, this technology can enable cheap, low-latency and low-power
tracking systems that  could enable
intrusive surveillance by malicious actors. Similarly, abuse
of technology in certain wearables is also possible. 

Again, we emphasize that it depends on the user to see the adaptation to either of these scenarios.

\section*{Acknowledgements}
We are grateful to Shikhar Jaiswal and Aayan Kumar for their
assistance in the deployment of \rpool models on Cortex-M4 devices. We
also thank Sahil Bhatia, Ali Farhadi, Sachin Goyal, Max Horton, Sham
Kakade and Ajay Manchepalli for helpful discussions and
feedback. Aditya Kusupati did a part of this work during his research
fellowship at Microsoft Research India.

\bibliography{local}
\clearpage
\appendix
\section{Dataset Information}
\label{sec:datasets}

\subsection{ImageNet-10}
\begin{wraptable}{r}{0.5\columnwidth}
\vspace{-8mm}
\centering
\caption{\small Classes in ImageNet-10 dataset.}
\begin{tabular}{@{}ccc@{}}
\toprule
Class no. & ImageNet id                      & Class name                                \\ \midrule
1         & {n02690373} & {`airliner'}         \\
2         & {n04285008} & {`sports car'}       \\
3         & {n01560419} & {`bulbul'}           \\
4         & {n02124075} & {`Egyptian cat'}     \\
5         & {n02430045} & {`deer'}             \\
6         & {n02099601} & {`golden retriever'} \\
7         & {n01641577} & {`bullfrog'}         \\
8         & {n03538406} & {`horse cart'}       \\
9         & {n03673027} & {`ocean liner'}      \\
10        & {n04467665} & {`trailer truck'}    \\ \bottomrule
\end{tabular}
\label{tab:imagenet-10-class}
\vspace{-6mm}
\end{wraptable} The ImageNet-10 is a subset of
images from ILSVRC 2012 ImageNet-1K
dataset~\citep{russakovsky2015imagenet} of 1000 classes. All images
corresponding to the 10 classes from CIFAR-10 as listed in
Table~\ref{tab:imagenet-10-class} are sampled from the full
dataset. The classes in CIFAR-10 are: airplane, automobile, bird, cat,
deer, dog, frog, horse, ship and truck.

The class n02430045: `deer' is not present in the ImageNet-1K subset
and was scraped from the full ImageNet-22K
database~\citep{deng2009imagenet}. Each class is divided into 1300
images for training and 50 images for validation.

Typical on-device models for real-world applications deal with limited
classes (e.g. intruder detection). ImageNet-10 is a good proxy for
this task with medium resolution natural images.

\subsection{Visual Wake Words}
This is a binary classification dataset~\citep{chowdhery2019visual}
dealing with the presence and absence of a person in the image. The
dataset is derived by re-labeling the images available in the MS COCO
dataset~\citep{lin2014microsoft} with labels corresponding to whether
a person is present or not. The training set has 115K images and the
validation set has 8K images. The labels are balanced between the two
classes: 47\% of the images in the training dataset of 115k images are
labeled as ‘person’.
 
\subsection{WIDER FACE}
This is a face detection dataset~\citep{yang2016wider} with 32,203
images containing 393,703 labeled faces varying in scale, pose, and
occlusion. It is organized based on 61 event classes. Each event class
has 40\%/10\%/50\% data as training, validation, and testing sets. The
images in the dataset are divided into Easy, Medium, and Hard
cases. The Hard case includes all the images of the dataset, and the
Easy and Medium cases are subsets of the Hard case. The hard case
includes images with a large number of faces or tiny faces along with
the data from Easy and Medium cases.

\subsection{SCUT HEAD}
This is a head detection dataset~\citep{peng2018detecting}. We use PartB
of this dataset for our experiments. PartB includes 2405 images with 43940
heads annotated. 1905 images of PartB are for training and 500 for testing.

\section{RNN as a spatial operator and comparison with ReNet}
\label{sec:renetapp}

Since ReNet~\cite{visin2015renet}, there have been a few methods that
have been built upon it to solve various vision tasks. The fundamental
difference, mathematically, between these approaches, and ours is how
the RNN is used to extract spatial information. In ReNet based
methods, the RNN is used to find a pixel-wise mapping from a voxel of
the input activation map to that of the output map. However, in our
method, we are using RNNs to spatially summarize a big patch of the
input activation map to a 1$\times$1 voxel of the output activation
map. Note that in ReNet the hidden states of every timestep of RNN
contribute to one voxel of the output, whereas in our case only the
last hidden states of the traversals are taken for both
row/column-wise summarizations and bidirectional summarizations.

ReNet based approaches either insert RNN based layers in existing
networks or replace a single convolution layer (thus resulting in
increasing computations). In ReNet, the RNNs are applied over the
whole input map, whereas RNNPool is applied patch by patch, which is
semantically similar to a pooling operator. Our usage of RNN for
spatial information extraction is so powerful that we can eliminate a
large amount of RAM and compute heavy convolution layers and still
preserve accuracy. For ReNet to do the same, patches of size equal to
the stride have to be flattened to construct an input to the RNN,
which makes it further inefficient in terms of compute and parameters
and results in loss of spatial dependencies. \rpool results in a
decrease in computations and parameters while ReNet based methods will
increase the same with respect to the baseline model. The comparisons
in Table \ref{tab:pooling} \& \ref{tab:imagenet} show that ReNet in
fact results in a significant loss in accuracy too.

\section{Probing the Efficacy of \rpool}
\subsection{Capturing Edges, Orientations and Shapes}
\label{sec:edges}

\begin{figure}[t]
    \centering
    \includegraphics[width=0.45\columnwidth]{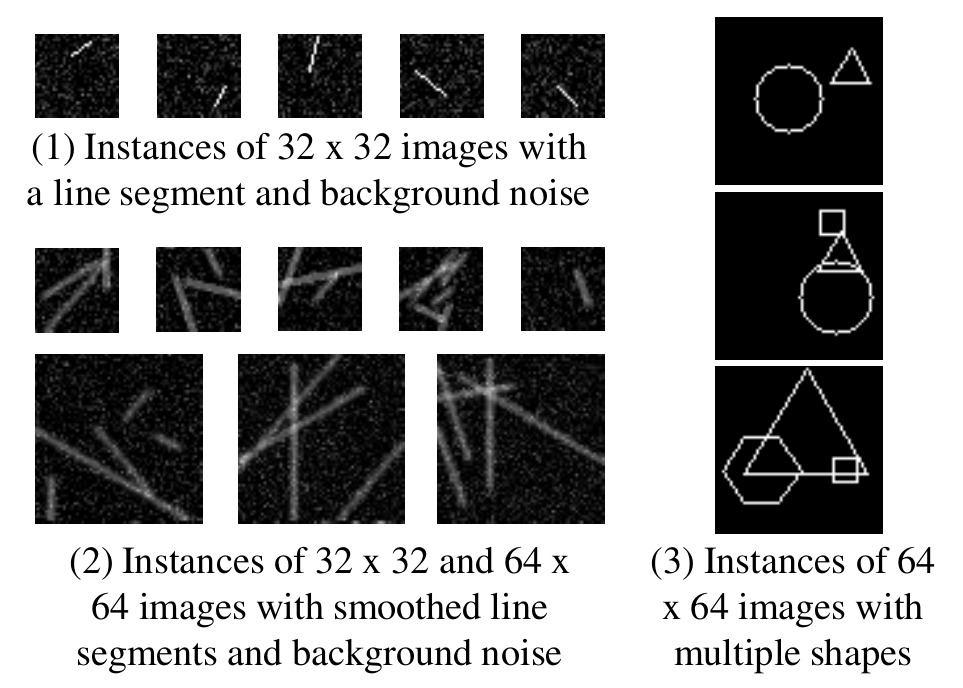}%
    \qquad
    \includegraphics[width=0.48\columnwidth]{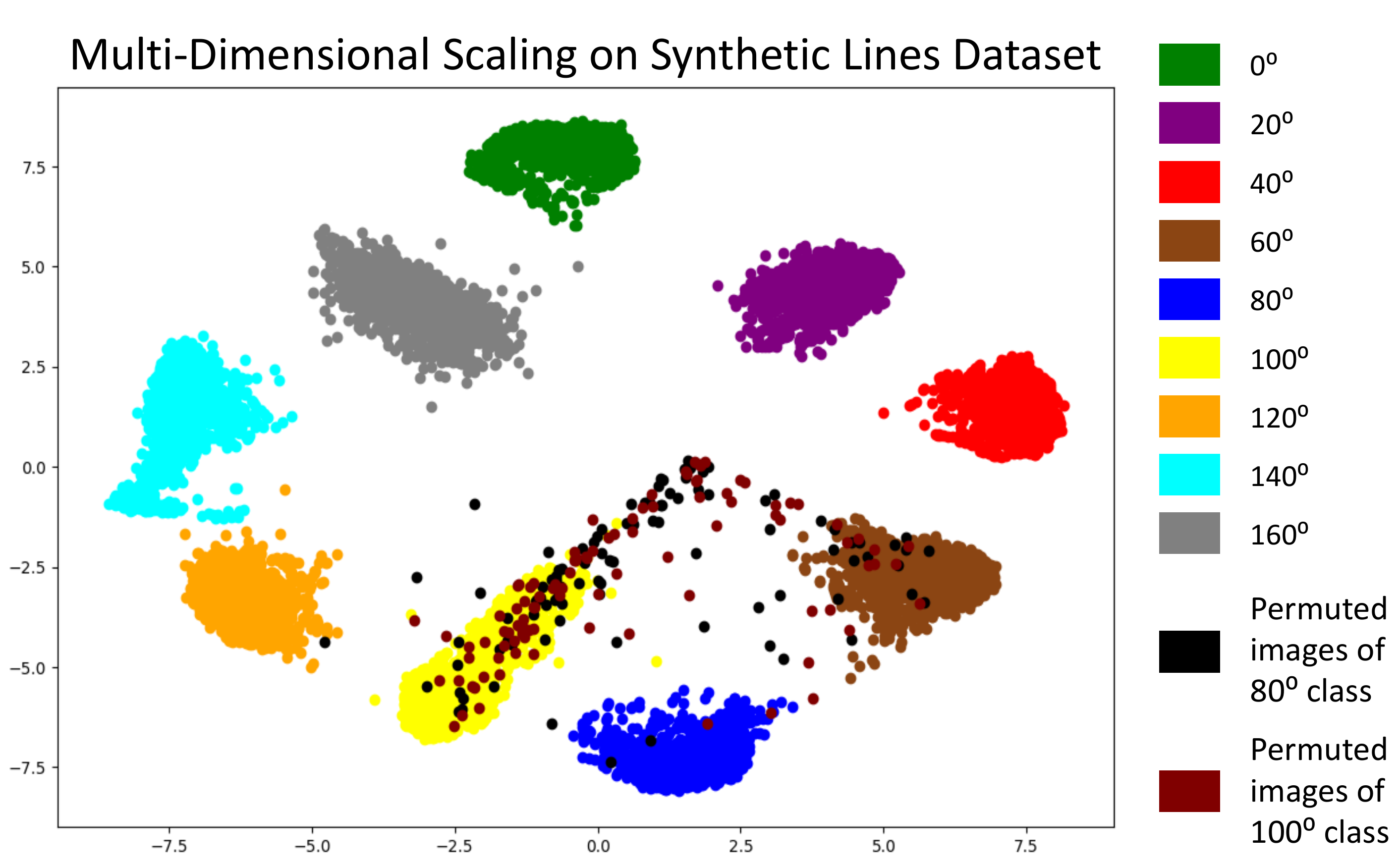}%
    \caption{(left) Examples from three multi-class and multi-label
      synthetic datasets used for probing \rpool. (right) A
      2-dimensional Multi-Dimensional Scaling visualization of the
      $128$ dimensional output of \rpool operator for the multi-class
      dataset (1). Some test images (plotted using black and brown
      dots) were modified by randomly permuting rows and columns.}%
    \label{fig:synthetic}%
\end{figure}

To probe \rpool's efficacy at capturing edges, orientation, and
shapes, we attempt to fit an \rpool operator to the following
synthetic datasets of small 8-bit monochrome images with background
noise as shown in Figure~\ref{fig:synthetic}.  We conduct experiments
on synthetic datasets to prove that \rpoollayer can learn spatial
representations. 
\begin{enumerate}[leftmargin=*]
  \item A multi-class dataset consisting of images with one line
    segment of varying lengths and positions. There are 9 classes
    corresponding to lines ranging from 0 to 160\textdegree~at
    20\textdegree~intervals.
  \item A multi-label dataset with images consisting of multizple line
    segments with varying lengths and positions. There are 9 labels
    corresponding to lines with orientations of 0 to 160\textdegree~at
    20\textdegree~intervals.
  \item A multi-label dataset consisting of images with a subset of
    shapes (5 in total) -- circle, triangle, square, pentagon, and
    hexagon.
\end{enumerate}

\begin{wraptable}{r}{0.5\columnwidth}
\vspace{-5mm}
\centering
    \caption{\small Minimum required hyperparameter configurations for synthetic experiments.}
\resizebox{0.5\columnwidth}{!}{%
\begin{tabular}{c c c c}
  \toprule
  Data  & Image Size &  With Conv. &  Without Conv. \\
  \midrule
  (1) & $32 \times 32$ & $h_1=4,h_2=16$  & $h_1=16,h_2=32$\\  
  (2) & $32 \times 32$ & $h_1=h_2=8$  & $h_1=h_2=32$\\  
  (2) & $64 \times 64$ & $h_1=8,h_2=16$  & $h_1=h_2=32$\\  
  (3) & $64 \times 64$ & $h_1=8=h_2=16$  & $h_1=h_2=32$\\  
  \bottomrule
\end{tabular}}
\label{tab:synconfig}
\vspace{-3mm}
\end{wraptable} We sweep over the $h_1, h_2$
parameters in powers of 2 for the smallest \rpool operator that can
enable a single FC layer to classify or label the test set with 100\%
accuracy. We do so with and without a preceding CNN layer of $8$
convolutions of $3\times 3$ size and stride
$2$. Table~\ref{tab:synconfig} lists the least $h_1, h_2$ required for
each task.  We observe that a single \rpool module fits to 100\%
accuracy on all these datasets.

We conclude that the horizontal and the vertical passes of the RNN
allows a single \rpool operator to capture the orientation of edges
and simple shapes over patches of size up to $64\times 64$.  Further,
adding a single convolutional layer before the \rpool layer makes the
model much more parameter efficient. In effect, the convolution layer
detects gradients in a local $3\times 3$ patch, while the \rpool
detects whether gradients across $3\times 3$ patches aggregate into a
target shape.

Further, we use multi-dimensional scaling~\citep{mead1992review} to
visualize the $4\cdot h_2 = 128$ dimensional output of \rpool operator
on the multi-class dataset (1) in Figure
\ref{fig:synthetic}~(left). Dataset (1) consists of various lines in
the image at a discrete set of angles, and the classification task is
to detect the angle of the line. Some images from the test set of
classes 80\textdegree~and 100\textdegree~are multiplied with a
permutation matrix to randomly permute rows and columns. These
resulting images are added to the original test dataset and the output
of the \rpool is plotted in Figure~\ref{fig:synthetic}~(right). The
outputs for each class form well-separated tight clusters indicating
\rpool indeed learns various orientations, while the outputs for the
permuted images are scattered across the plot indicating that it is
not exploiting certain gross aggregations in the data.

\subsection{Comparing Performance with Pooling Operators}
\label{sec:cifar}
We now contrast the down-sampling power of \rpool against standard
pooling operators. That is, we investigate if the pooling operators
maintain accuracy for a downstream task even when the pooling
receptive field is large. To this end, we consider the image
classification task with CIFAR-10 dataset~\citep{krizhevsky2009learning} but the pooling operator is
required to down-sample the input $32\times32$ image to a $1\times1$
voxel in {\em one go } i.e. both patch size and stride are 32. This is
followed by a fully connected (FC) layer. The number of output
channels after pooling was ensured to be the same. For Max and Average
pooling models, a $1\times 1$ convolution is used to ensure the same
output dimension. For this task, \rpool achieves an accuracy of
{$\textbf{70.63\%}$}, while the convolution layer, max pooling, and average
pooling's accuracy are $53.13\%$, $20.04\%$ and $26.53\%$,
respectively. This demonstrates the modeling power of the \rpool
operator over other pooling methods. Table~\ref{tab:pooling}~(Rows
2-5) reinforces the same but on bigger image classification datasets.

\textbf{Details}. We use $h_1=h_2=32$ for the \rpool operator with
patch size and stride as 32. For the strided convolution we use a
convolution layer of $4 \times h_2=128$ filters. For Max and Average
pooling first we pool down to $1 \times 1 \times 3$ from input of $32
\times 32 \times 3$ and then use a $1 \times 1$ convolution of 128
filters. All the above have the same patch size and stride size and
are followed by a fully connected layer projection to 10 from 128.



\section{Lower bounds on space required for multi-layer networks}
\label{sec:space}

We now lower bound the memory requirements of computation of
multi-layer convolutional networks when recomputation is not
permitted.  Suppose we have an $l$-layer ($l>1$) convolutional network.
Let $Y$ denote the nodes in the final layer which form a grid of size
$m\times n$.  Suppose that the size of the receptive field of each node in
$Y$ in an intermediate layer $l$ is $(2k+1)\times(2k+1), k>0$ and that
$y_{i,j}\in Y$ depends on the activations of nodes $x^{(l)}_{i'j'},
i'\in\{i-k, \dots, i, \dots i+k\}, j'\in\{j-k, \dots, j, \dots j+k\}$ in the
intermediate layer $l$.  Suppose further that the convolution
operations have stride $1$ and are generic and not separable, i.e.,
can not in the general case be factored into depth-wise separable
operations.  An execution of this network ``disallows recomputation''
if once a node $x$ in an intermediate layer (layers that are neither
the input nor output of the network) is computed, all nodes $y\in Y$
that depend on $x$ must be computed before $x$ is evicted from memory.

\begin{claim}
\label{cl:mem}
Fix column $j\in [n]$. Suppose that nodes $y_{i,j},\ i \in I\subsetneq
[m]$ have been completed at some point in an execution order. Then at
the same point in the execution order, at least $2k$ contiguous
activations $x^{(l)}_{i^{*}-k+1,j}, x^{(l)}_{i^{*}-k+2,j}, \dots
x^{(l)}_{i^{*}+k,j}$ for some $i^{*}\in[m]$ will need to be saved in
memory until another node from column $j$ is computed.
\end{claim}

\textbf{Proof.}
Since $I\subsetneq[m]$, there exists index $i^{*}\in[m]\setminus I$
such that either $i^{*}+1 \in I$ or $i^{*}-1 \in I$. Suppose without
loss of generality that $i^{*}-1 \in I$. Then, nodes
$x^{(l)}_{i^{*}-k+1,j}, x^{(l)}_{i^{*}-k+2,j}, \dots,
x^{(l)}_{i^{*}+k-1,j}$ must have been loaded into memory. However,
$y_{i^{*},j}$ also depends on these intermediate nodes, and has not
yet been computed. So these $2k$ intermediate nodes must be retained
in memory, thus proving the statement. The case where $i^{*}+1 \in I$
is similar.

With this claim, we are ready to prove Proposition~\ref{prop:mem}.

\textbf{Proof of Proposition~\ref{prop:mem}.}
Fix any execution order of the network, and label the nodes in the
final layer $Y$ in the order they are evaluated: $(p_1, q_1), (p_2,
q_2), \dots, (p_{mn}, q_{mn})$. That is $y_{p_1, q_1}$ is evaluated
before $y_{p_2, q_2}$ and so on. Let
\[
I_t = \cup_{\tau=1}^{t} p_\tau,\quad
J_t = \cup_{\tau=1}^{t} q_\tau,\quad
\textrm{and}\quad
t^{*} =\min_t \{|I_t|=m \ \textrm{or}\ |J_t|=n\}.
\]
That is, once $y_{p_{t^{*}}, q_{t^{*}}}$ is executed, either (a) at
least one node in each row of the final layer has been executed, or
(b) at least one node in each column of the final layer has been
executed, and at the moment $y_{p_{t^{*}-1}, q_{t^{*}-1}}$ is
computed, there is an entire row, say $r$, and an entire column, say
$c$, in the final layer where no nodes have been executed.

Suppose that case (b) holds. Then, at step $t^{*}-1$, nodes in $n-1$
columns $[n]\setminus \{c\}$ have been executed, and in each column,
at least one row has not been executed. By Claim~\ref{cl:mem}, each such column
would need to have $2k_q$ activations at layer $q$ in memory at this point
of execution, and all these nodes are unique (that the nodes required
to be in memory by Claim~\ref{cl:mem} for different columns are non-overlapping).
Therefore, at least $2\sum_{q=1}^{l-1} c_q k_q \times (n-1)$ memory is required
to hold the necessary nodes in each intermediate layer for this execution.

A similar analysis of case (a) yields a lower bound of
$2\sum_{q=1}^{l-1} c_q k_q \times (m-1)$ from which the lemma follows.
\hfill $\blacksquare$

If convolution operators have a stride larger than $1$, then we can
similarly state the following claim based on the overlap between the
nodes in an intermediate layer that are common dependencies across two
consecutive rows/columns of the output.

\begin{claim}
\label{cl:mem-stride}
Fix column $j\in [n]$. Suppose that nodes $y_{i,j},\ i \in I\subsetneq
[m]$ have been completed at some point in an execution order. Suppose
that the stride at layer $q$ is $s_q$. Restrict $s_q$ to $1$ in a
layer with $1\times 1$ convolutions, i.e., assume activations are not
simply thrown away. Then at the same point in the execution order, at
least $k'=2k+1- \Pi_{r=q}^{l} s_r$ contiguous activations
$x^{(l)}_{i^{*}-\lfloor k'/2\rfloor +1 ,j}, x^{(l)}_{i^{*}-k+2,j},
\dots x^{(l)}_{i^{*}+\lceil k'/2\rceil,j}$ for some $i^{*}\in[m]$ will
need to be saved in memory until another node from column $j$ is
computed.
\end{claim}

This allows us to restate Proposition~\ref{prop:mem} in networks
where stride is greater than $1$.

\begin{proposition}
	\label{prop:mem-stride}
	Consider an $l$-layer ($l>1$) convolutional network with a
        final layer of size $m\times n$.  Suppose the for each node in
        the output layer, the size of receptive field in intermediate
        layer $q \in[l-1]$ is $(2k_q+1)\times(2k_q+1), k_q>0$ and that
        this layer has $c_q$ channels and stride $s_q$.  Restrict
        $s_q$ to $1$ in a layer with $1\times 1$ convolutions. Suppose
        that $k'_q = 2k_q +1 - \Pi_{r=q}^{l-1} s_r$.  Any serial
        execution order of this network that disallows re-computation
        requires at least $\sum_{q=1}^{l-1} c_q k'_q \times
        min((\Pi_{r=q}^{l-1} s_r) m -1 , (\Pi_{r=q}^{l-1} s_r)n-1)$
        memory for nodes in the intermediate layers.
\end{proposition}

\begin{claim}
\label{cl:upper-interm}
The lower bound in Proposition~\ref{prop:mem} is matched by an
execution order that computes the network in a row or column-first
order, whichever is smaller. That is, execute all the intermediate
nodes needed to compute the first row of the output, retain those
intermediate nodes required for the calculation of the second row of
the output, compute the second row of output, and so on.  Let $S_q =
\Pi_{r=q}^{l} s_r$, and restrict $s_q$ to $1$ in a layer with $1\times
1$ convolutions.  This schedule has a memory requirement of $
\sum_{q=1}^{l-1} c_q (2k_q+1 - S_q) min(Q_q m-1 + 2k_q, S_q n-1+2k_q)$
if we account for the padding at either ends of the row in each
intermediate layer, and
\[
\sum_{q=1}^{l-1} c_q (2k_q+1 - S_q) min(S_{q} m-1, S_{q} n-1),
\]
if the padding is not counted.
\end{claim}

\begin{claim}
\label{cl:upper}
Suppose we follow the row (or column)-wise execution order in
Claim~\ref{cl:upper-interm}, and that each row in the output depends
on $k_0$ layers at the input. Suppose that the input is required to be
in memory before the start of the execution and the output is required
to be in memory at the end of the execution. Let $c_{in}$ and
$c_{out}$ denote the number of channels in the input and output.  Let
$S_q = \Pi_{r=q}^l s_r$, and let $k'_0 = k_0 - S_1$ be the number
of rows/columns in the input layer that are common dependencies
between two consecutive rows/columns of the output. The memory
requirement including \textbf{those of the input and output layers} is
\[
\max\{m_{in}n_{in}c_{in} + k'_0 n_{out}c_{out}, m_{out}n_{out}c_{out} +
k'_0n_{in}c_{in} \} + \sum_{q=1}^{l-1} c_q (2k_q+1 -S_q) min(S_q m-1,
S_q n-1),
\] with
padding added on the fly for convolutions at the boundaries of
activation maps. This is obtained by reclaiming the footprint of the
input for the output one row at time (with a lag of $k_0$ rows) once
all the nodes that depend on it are completed.
\end{claim}

\section{Details about Compute and Peak RAM Calculation}
\label{sec:peak}

In this section, we quantify the memory requirements of the networks
analyzed in this paper.

\subsection{Optimal memory requirements  without recomputation}
\label{sec:opt-mem}
First, we analyze the minimum memory requirements and optimal
execution orders of components -- inverted residual block, separable
residual block, dense block, and inception block -- assuming that no
re-computation is allowed.  That is, we wish to find the minimum
value, over all valid execution orders $E$ of the block, of the
maximum memory requirement of the execution order.  Then, we analyze
the memory requirement of image classification architectures discussed
in this paper.

\subsubsection{Memory requirements of various block}
We assume that the
execution always starts with the input of the block in memory, and
terminates with output in memory. We denote that the size of input $I$
is $h_{in}\times w_{in} \times C$, where $h_{in}$ and $w_{in}$ are the
height and the width of the activation and $c_{in}$ is the number of
channels. Likewise, denote the size of $O$ to be $h_{out} \times
w_{out} \times c_{out}$. In what follows, suppose also that $h_{in}
\geq w_{in}$ and $h_{out} \geq w_{out}$. Otherwise we can flip rows
and columns and meet the same constraints.

\begin{enumerate}[leftmargin=*]
  \item \textbf{Inverted bottleneck residual block (a.k.a. MBConv, see
    Fig. 3b of~\citep{sandler2018mobilenetv2})} : The first layer is a
    point-wise convolution (C1) that expands the number of channels to
    $c_{in} \times t$ where $t$ is expansion factor. Then there is a
    depth-wise separable $3 \times 3$ convolution (C2) with stride
    either 1 or 2, followed by another point-wise convolution (C3)
    which reduces the number of output channels.  We can use the
    row-wise order suggested in Claim~\ref{cl:upper}, which results in
    a schedule where the first row of the output is generated, then
    the second row and so on. This schedule has a memory footprint of
    $\max\{h_{in}w_{in}c_{in} + (3-s)w_{out}c_{out},
    h_{out}w_{out}c_{out} + (3-s)w_{in}c_{in}\} + (3-s)tc_{in}w_{in}
    $, where $s$ is the stride of the $3\times 3$ convolution.


  \item \textbf{Residual Block (see Fig. 5(left)
    of~\citep{he2016deep})} : We consider a residual block consisting
    of two convolution layers with $3\times 3$ kernels, of which the
    first has a stride $s$ of 1 or 2, and the second has stride $1$.
    The we have $w_{out} = w_{in}/s$ and $h_{out} = h_{in}/s$.  Using
    Claim~\ref{cl:upper}, we can see that the best case memory
    footprint is $\max\{h_{in}w_{in}c_{in} + (5-s)w_{in}c_{out}/s,
    h_{in}w_{in}c_{out}/s^2 + (5-s)w_{in}c_{in}\} + 2w_{in}c_{out}/s$,
    assuming that the number of channels of intermediate layer is
    equal to $c_{out}$ as is the norm here.

  \item \textbf{Inception block (see Fig. 2b
    of~\citep{szegedy2015going})}: Denote the output of each of the
    $4$ paths in the block by $O_1, O_2, O_3$ and $O_4$. We consider
    the case where all convolutions are of stride $1$. We can apply
    the arguments of Section~\ref{sec:space} simultaneously for all
    four paths with slight modification.  We consider a minimal set of
    contiguous rows at the start of the input -- which would be first
    5 row in the referenced image as its the largest convolution size
    -- and compute all channels in the first row of the output of all
    four paths.  We then drop the first row of input, materialize the
    second row of output on all four paths and so on. If we denote by
    $c_{out}$ the number of output channels of all four networks, then
    the memory requirement is $\max\{h_{in}w_{in}c_{in} + 4
    w_{out}c_{out}, h_{out}w_{out}c_{out} + 4w_{in}c_{in}\} + (2c_2 +
    4c_3)w_{in}$, where $c_2$ and $c_3$ are the number of intermediate
    channels in $O_2$ and $O_3$ respectively.


  \item \textbf{Dense block (see Fig. 4 of
    \href{https://towardsdatascience.com/understanding-and-visualizing-densenets-7f688092391a?gi=94436891b97}{URL})}
    : At any point in the execution of a dense block, we need to store the input to the dense block and outputs of all previous dense layers, since the last layer needs all the activation maps concatenated as its input. The total activation maps being stored will reach the peak just after the last dense layer. Therefore the peak memory requirement is the output of the dense block.
\end{enumerate}

\subsubsection{Memory requirements of image classification networks}

\begin{table*}[t]
  \centering
  \caption{\small Comparison of accuracy, compute and minimum memory
    requirement for inference with and without \rpoollayer on
    ImageNet-10. The memory calculations reflect the application of
    Proposition~\ref{prop:mem-stride} and Claim \ref{cl:upper}}.
  
  \resizebox{\columnwidth}{!}{%
    \begin{tabular}{ l | c  c  c  c | c  c  c  c }
      \toprule
      \multicolumn{1}{l|}{\multirow{2}*{Model}} &  \multicolumn{4}{|c|}{Base} & \multicolumn{4}{|c}{{\alg}} \\\cmidrule{2-9}
      & Accuracy (\%) & Parameters & Peak RAM  & MAdds & Accuracy (\%) & Parameters & Peak RAM  & MAdds \\
      \midrule
      MobileNetV2 & 94.20 & ~2.20M & 0.84MB & 0.30G & \textbf{94.40} & \textbf{~2.00M} & \textbf{0.24MB} & \textbf{0.23G} \\
      EfficientNet-B0 & 96.00 & ~4.03M & 0.84MB & 0.39G & \textbf{96.40} & \textbf{~3.90M} & \textbf{0.24MB} & \textbf{0.33G}\\
      ResNet18 & \textbf{94.80} & 11.20M & 0.81MB & 1.80G & 94.40 & \textbf{10.60M} & \textbf{0.38MB} & \textbf{0.95G} \\
      DenseNet121 & \textbf{95.40} & ~6.96M & 2.38MB & 2.83G & 94.80 & \textbf{~5.60M} & \textbf{0.77MB} & \textbf{1.04G} \\
      GoogLeNet & \textbf{96.00} & ~9.96M & 1.01MB & 1.57G & 95.60 & \textbf{~9.35M} & \textbf{0.59MB} &  \textbf{0.81G} \\
      \bottomrule
   \end{tabular}}
\vspace{-5mm}
  \label{tab:10classramoptimized}
\end{table*}


We calculate the lowest possible memory requirements of networks using
calculations in the previous subsection for individual blocks and the
following methodology: find a partitioning of a multi-layer network
into disjoint contiguous sets of layers that minimizes the least
memory requirement of the most memory-intensive partition.  Using
this, we calculate the memory requirements of networks in Table
\ref{tab:10class} and list the requirements in
Table~\ref{tab:10classramoptimized}. We now discuss the specifics of
each network, and in particular, the partition of the layers of the
network that requires the maximum memory (and thus lower bonds the
memory requirement of a network).

\textbf{GoogLeNet} has a initial convolution layer (C1) of stride 2,
followed by a max pooling layer (P1), another convolution layer (C2)
of stride 2 and then a max pooling layer (P2). Output of P2 is of size
$28 \times 28 \times 192$. Applying
Proposition~\ref{prop:mem-stride} to the set of layers starting with
the input image ($I$) and output of P2 ($O$), the RAM required is 112
$\times$ (11-4) $\times$ 64 + 56 $\times$ (5-2) $\times$ 64 + 56
$\times$ (3-2) $\times$ 192 added to $O$ and 7 rows of input, is
lesser than the requirement for inception (3b). For the inception (3b)
block, the input is ( 28 $\times$ 28 $\times$ 256) and the output is
of size 14 $\times$ 14 $\times$ 480. Therefore using
Proposition~\ref{prop:mem-stride}, the RAM required is 28 $\times$
(7-2) $\times$ 32 + 28 $\times$ (5-2) $\times$ 128 + 28 $\times$ (3-2)
$\times$ 64 + 28 $\times$ (3-2) $\times$ 480 (the first three terms
are intermediate activations of the inception block and have different
receptive fields), added to the input size (28 $\times$ 28 $\times$
256) + 14 $\times$ (7-2) $\times$ 480, results in 1.01MB.

\textbf{DenseNet121} has a 2-strided convolution layer (C1) in the
beginning followed by a max pool of stride 2 (P1) and then D1-the
first Dense block which has 6 Dense layers. Each Dense layer has
$1\times1$ convolution with 128 output channels followed by a
$3\times3$ convolution with 128 input and 32 output channels. The
output of each Dense layer is concatenated to the input to form the
input to the next Dense layer which is why the $1\times1$ convolution
in each Dense layer has different input channels. D1 is followed by a
$1\times1$ convolution which reduces channels of activation map to
half followed by P2, another Max Pool layer. For determining the peak
RAM required, we apply Proposition~\ref{prop:mem-stride} to the set of
layers starting with the output of P1 ($I$) until the output of P2
($O$), so that we can go from 56 $\times$ 56 $\times$ 64 to 28
$\times$ 28 $\times$ 128 directly bypassing 56 $\times$ 56 $\times$
256 sized $O_{D1}$. The receptive field of $O$ on $I$ can be
calculated to be 14$\times$14. The RAM for intermediate activations
will be 56 $\times$ (14-2) $\times$ 128 + 56 $\times$ (12-2) $\times$
32 + 56 $\times$ (12-2) $\times$ 128 + 56 $\times$ (10-2) $\times$ 32
+ \dots + 56 $\times$ (4-2) $\times$ 32. The total peak RAM along with
$I$ (56$ \times$ 56 $\times$ 64) + 28 $\times$ (14-2) $\times$ 128,
which is 2.38MB.



\textbf{ResNet18}. A similar calculation as above can be done for
ResNet18. The architecture consists of a convolution layer (C1) of
stride 2 followed by a max pool layer (P1), followed by residual
blocks. In this case, let us apply Proposition~\ref{prop:mem-stride}
to the block of layers starting with the input RGB image of size 224
$\times$ 224 $\times$ 3 (denoted $I$) until the output of P1 (denoted
$O$). Between $I$ and $O$ we have 2 layers: C1 and P1. Therefore the
total RAM requirement will be 112 $\times$ (3-2) $\times$ 64 added to
$O$ (56 $\times$ 56 $\times$ 64) + 224 $\times$ (11-4) $\times$ 3,
which is 0.81MB.


\textbf{MobileNetV2} has a convolution layer C1 of stride 2 followed
by a MBConv block MB1 which has stride 1. MB1 contributes to the peak
memory (2.29MB). Denote by $I$ the input RGB image of size 224
$\times$ 224 $\times$ 3 and denote by $O$ the output of MB1. The
receptive field of $O$ on output of C1 is 3, on output of first layer
of MB1 is 3 and after the 1 for the rest two layers of MB1. Therefore,
using Proposition~\ref{prop:mem}, the RAM required is 112 $\times$
(3-1) $\times$ 32 + 112 $\times$ (3-1) $\times$ 32 added to $O$ ( 112
$\times$ 112 $\times$ 16 )) + 224 $\times$ (7-2) $\times$ 3, which is
0.84MB.

\textbf{EfficientNet-B0} has exactly the same calculation as
MobileNetV2 as the first convolution block and first MBConv block are
identical.

\textbf{RNNPool Versions} : Similar to GoogLeNet we can also reduce
peak RAM of GoogLeNet-\rpool. Here inception (4e) is the bottleneck.
Lets take $I$ as the input to inception (3b)( 14 $\times$ 14 $\times$
528) and $O$ as the output of the pooling layer after inception
(3b). Size of $O$ is 7 $\times$ 7 $\times$ 832. Therefore using
Proposition~\ref{prop:mem}, the RAM required is 14 $\times$ (7-2)
$\times$ 32 + 14 $\times$ (5-2) $\times$ 160 + 14 $\times$ (3-2)
$\times$ 128 + 14 $\times$ (3-2) $\times$ 832, added to input (14
$\times$ 14 $\times$ 528) + 7 $\times$ (7-2) $\times$ 832, resulting
in 0.59MB.

The peak memory requirements of \rpool versions of ResNet18,
DenseNet121, MobileNetV2 and EfficientNet-B0 in
Table~\ref{tab:10class} cannot be reduced further by better schedules
as we replace the most memory-intensive blocks and operate
patch-by-patch, which is more local and granular that row-by-row
schedules used above.


\subsection{Memory requirement (without recomputation) estimates according to prior conventions}
\label{sec:compute-opt}


In this subsection, we follow the scheduling convention
of~\citet{chowdhery2019visual} to estimate the memory requirements of
individual blocks and networks that use them. Note that the memory
requirements listed here can be higher than in
Section~\ref{sec:opt-mem} as the schedules may not be optimal from
memory requirement perspective.

\subsubsection{Memory requirements of individual blocks}

\begin{enumerate}
  \item \textbf{Inverted bottleneck residual block (a.k.a. MBConv)} :
    Give input $I$ of size $h_{in}\times w_{in} \times C$, a pointwise
    convolution (C1) first expands the number of channels to $C \times
    t$ where $t$ is expansion factor. Then there is a depthwise
    separable $3 \times 3$ convolution (C2) with stride either 1 or 2,
    followed by another pointwise convolution (C3) which reduces the
    channel to the number of output channels ($O$) associated with the
    MBConv block. To avoid storing the large output ($O_{C1}$) of C1
    and bloating the memory, $O_{C1}$ is constructed channel by
    channel, so at first 1 filter of the $C \times t$ filters of C1
    will be convolved with $I$, then this single 2D vector will be
    convolved by C2. Since C2 is depthwise separable and input
    channels independently contribute to an output channel, we again
    get a 2D map. This map is convolved with all filters of C3 and we
    get an output of $O$ number of channels. We keep doing this, going
    one by one through each filter of C1 and adding to the output of
    the MBConv block of $O$ channels, to get the final output. Hence,
    the memory requirement is the size of input added to that of the
    output of the MBConv block.
  
  \item \textbf{Residual Block} : The memory requirement is the
    maximum of input and output maps of the block. As the residual connection adds the input to the output values can be discarded after being added to the output values being computed.
  
  \item \textbf{Inception block}: Denote the input to the inception
    block $I$ and the outputs of each of the $4$ paths in the block
    $O_1, O_2, O_3$ and $O_4$. Since we can get rid of the input $I$
    after computing the last output, we can order the computation in increasing order of the number of channels in $O_i$. Therefore,
    the peak RAM while computing the full block will be the sum of input added to the sum of the 3 smallest outputs.
  
  \item \textbf{Dense block}: A dense block needs to store the input as well as outputs of all previous dense layers since the last layer needs all the activation maps concatenated. The volume activation maps stored will reach the peak just after the last dense layer. Therefore the peak RAM usage is the size of the output of the dense block.
\end{enumerate}

\subsubsection{Memory requirements of image classification networks in Table \ref{tab:10class}}
\label{sec:ramcalculation10class}
We now use the above results to compute the memory requirements of
image classification networks, assuming all computations are in 32-bit
floating-point. We assume the layer-by-layer convention of
~\cite{chowdhery2019visual} for RAM computation.  The peak memory
requirement of both MobileNetV2 and EfficientNet-B0 is contributed by
the first MBConv block in these architectures. The input map size to
the block is $112 \times 112 \times 32$ and the output map size is $112
\times 112 \times 16$, adding up to a peak memory requirement of 2.29MB.

The peak memory requirement of the \rpool inserted versions is the
MBConv block right after the \rpool replacement. The input size is $28
\times 28 \times 64$ and output size is $14 \times 14 \times 64$ for
MobileNetV2-\rpool, adding up to 0.24MB. The input size is $28 \times
28 \times 64$ and output size is $14 \times 14 \times 80$ for
EfficientNetB0-\rpool, adding up to 0.25MB.

For ResNet18, DenseNet121, and GoogLeNet the maximum memory requirement
is to host the activation map just after the first convolution layer
which is of size $112 \times 112 \times 64$.  For ResNet18-\rpool, the
maximum requirement comes from the residual block just after \rpool,
i.e., the first residual block out of the two of conv4\_x. The input
to this is of size $28 \times 28 \times 128$ and the output size is $14
\times 14 \times 256$. The maximum of these two is 0.38MB. For
DenseNet121-\rpool, the largest memory requirement comes from the
output of D3 (see Figure \ref{fig:DenseNet-PRNN}), the size of which
$14 \times 14 \times 1024$ i.e. 0.77MB. For GoogLeNet, the peak
requirement comes from the last inception block on the spatial resolution
of $14 \times 14$ --- inception (4e).  Here the size of the input is $14
\times 14 \times 528$ and sizes of the 3 smallest outputs are $14
\times 14 \times 128$, $14 \times 14 \times 128$ and $14 \times 14
\times 256$, totaling 0.78MB.

\subsubsection{Memory requirement of  face detection networks in Table \ref{tab:facedetection} without recomputation}
We use convention of considering the largest activation map to be the
peak RAM requirement. For EagleEye, FaceBoxes, EXTD and LFFD
architectures, the largest activation map is the output of the first
convolution, their sizes being $320 \times 240 \times 4$ (=1.17MB),
$160 \times 120 \times 24$ (=1.76MB), $320 \times 240 \times 64$
(=18.75MB) and $320 \times 240 \times 64$ (=18.75MB) respectively. For
\rpool-Face-A and \rpool-Face-B, the largest activation map is the
output of the \rpool, which is $160 \times 120 \times 16$ (=1.17MB)
and $160 \times 120 \times 24$ (=1.76MB) respectively.  For
\rpool-Face-C and \rpool-Face-Quant, peak memory requirement is
contributed by the MBConv block right after the \rpool. The input size
of this block for \rpool-Face-C is $160 \times 120 \times 64$ and
output size is $160 \times 120 \times 24$, the total being 6.44MB. The
input size of this block for \rpool-Face-Quant is $80 \times 60 \times
32$ and output size is $80 \times 60 \times 16$, the total being 224KB
as we quantize to 1 byte unsigned integer.

\subsection{Memory requirements of image classification networks in Table \ref{tab:10class} with recomputation}
\label{sec:with-recomp}

As explained in Section \ref{sec:ramcalculation10class}, the RAM
calculations for \rpool based models revealed that the convolution
block after \rpoollayer contributes to the peak RAM. Let's denote this
block in both the base architecture and \rpool-based version as
ConvBlock-A. In the memory-optimized scheme, we fix the peak RAM of
the base model to be that of the convolution block whose RAM usage is
a bit more than that of the \rpool version. We denote by ConvBlock-B the
convolution block that lies before ConvBlock-A, and such that there
exists no block that lies between this block and ConvBlock-A which has
a RAM usage less than that of ConvBlock-A. Note that ConvBlock-B is
present only in the base model and not the \rpool model. Since we fix
the peak RAM, we have to reconstruct an activation map (denoted by
Activation-A) that comes before ConvBlock-B patch by patch. Note that Activation-A
need not necessarily be the activation map just before
ConvBlock-B. Activation-A is chosen as the earliest occurring
activation map (nearer to the input image) which ensures that there is no
intermediate layer or block between it and ConvBlock-B which can
contribute to more RAM usage. We do construct Activation-A
by loading a patch of the image (one at a time), which is of the size
of the receptive field of Activation-A w.r.t. the input image, and
feed it forward to get a $1 \times 1 \times channel_{Activation-A}$
voxel of Activation-A. When we load the next patch we have to
re-compute some convolution and pooling outputs which come in the
overlapping region of the two consecutive patches. We keep doing this
until we reconstruct Activation-A completely. The total number of MAdds is the sum of the MAdds of the base network and the extra re-computations in order to
compute patch-by-patch.

\section{Architectures}
\label{sec:architectures}
\subsection{Image Classification}
\label{sec:arch-classification}

\subsubsection{\rpoollayer in the beginning replacing multiple blocks}
\begin{table}[!ht]
\centering
    \caption{\small \rpool settings for image classification.}

\begin{tabular}{l | c |c}
  \toprule
  Model  & Hidden Size &  Patch Size \\
  \midrule
  MobileNetV2-\rpool & $h_1=h_2=16$ & 6\\  
  EfficientNet-B0-\rpool & $h_1=h_2=16$ & 6\\  
  ResNet18-\rpool & $h_1=h_2=32$ & 8\\  
  DenseNet121-\rpool & $h_1=h_2=48$ & 8\\
  GoogLeNet-\rpool & $h_1=h_2=32$ & 8\\
  MobileNetV2-\rpool (0.35$\times$) & $h_1=h_2=8$ & 6\\
  \bottomrule
\end{tabular}
\label{tab:rpool-hparam-img}
\end{table}
\label{sec:toprep}
As discussed in Figure~\ref{fig:DenseNet-PRNN}, we can use \rpoollayer in the beginning of the architecture to rapidly downsample the image leading to smaller working RAM and compute requirement. Table~\ref{tab:rpool-hparam-img} presents the hidden state size and patch size used by \rpoollayer when applied to various models discussed in Table 
\ref{tab:10class}. 
Note that the last row refers to the model used for Visual Wake Words experiments (Figure~\ref{fig:vww}).

Furthermore, Table~\ref{tab:mobilenetv2-rp} presents the exact architecture used by MobileNet-v2-RNNPool(0.35x) architecture applied to the Visual Wakeword problem (Section~\ref{sec:rp+vww}).  
\begin{table}[h]
\centering
    \caption{\footnotesize MobileNetV2-\alg : \rpool Block with patch-size 6$\times$6 and hidden sizes $h_1=h_2=16$ is used.
    The rest of the layers are defined as in \cite{sandler2018mobilenetv2}. Each line denotes a sequence of
    layers, repeated $n$ times. The first layer of each bottleneck
    sequence has stride $s$ and rest use stride $1$. Expansion factor
    $t$ is multiplied to the input channels to change the width. The number of output classes is $l$.}
    \begin{tabular}{ c | c | c | c | c | c }
      \toprule
	  {Input} & {Operator} & $t$ & $c$ & $n$ & $s$\\
	  \midrule
	  $224^2 \times 3$ & conv2d $3\times3$ & 1 & 32 & 1 & 2\\
	  $112^2 \times 32$ & \alg Block & 1 & 64 & 1 & 4\\
	  $28^2 \times 64$ & bottleneck & 6 & 64 & 4 & 2\\
	  $14^2 \times 64$ & bottleneck & 6 & 96 & 3 & 1\\
	  $14^2 \times 96$ & bottleneck & 6 & 160 & 3 & 2\\
		$7^2 \times 160$ & bottleneck & 6 & 320 & 1 & 1\\
	  $7^2 \times 320$ & conv2d $1\times 1$ & 1 & 1280 & 1 & 1\\
	  $7^2 \times 1280$ & avgpool $7\times 7$ & 1 & - & 1 & 1\\
	  $1 \times 1 \times 1280$ & conv2d $1\times 1$ & 1 & $l$ & - & 1\\
	  \bottomrule
  \end{tabular}
    \label{tab:mobilenetv2-rp}
\end{table}

\subsubsection{\rpoollayer replacing Average Pooling at the end}
Typical image classification models use average pooling before the final feed-forward layer to produce the class probabilities. As \rpoollayer is syntactically equivalent to standard pooling layers, we can use it to perform the pooling in the penultimate layer, replacing the average pool layer. To this end, we use \rpool operator with  $h_1=h_2=l/4$ where $l$ is the number of channels in the last activation map before the average pooling layer. Such a replacement does not significantly contribute to the number of parameters and MAdds. In Table \ref{tab:pooling}, Row 2 refers to such a replacement in the base MobilnetV2, DenseNet121, and MobilenetV2-0.35x models, while Row 7 refers to similar replacement in the corresponding \rpool models. In Figure \ref{fig:vww}, all \rpool based architectures use \rpool both in the beginning layer and in the penultimate layer of the network. 

\subsubsection{\rpoollayer replacing intermediate Pooling layers}

These experiments have been tried on DenseNet121 as the base model (Section-\ref{sec:usage}), where we are replacing single max-pooling layers appearing in intermediate positions in the network with \rpool. Given $r_{in} \times c_{in} \times k_{in}$ size input activation map to the pooling layer, the hidden sizes for \rpool is taken as $h_1=h_2=k_{in}/4$, patch size as 4 and stride as 2. Note that we also further drop dense layers ($1 \times 1$ convolution followed by $3 \times 3$ convolution) in D3 and D4. The number of channels in the output of any dense block is the sum of the number of input channels and output of each dense layer. Hence, reducing the number of dense layers reduces the number of channels of the output activation maps of these dense blocks and hence the input to the pooling layer. However, for the \rpool the same strategy of $h_1=h_2=k_{in}/4$ is followed where $k_{in}$ is lesser now.


\subsection{Face Detection}
\label{sec:fdmodels}

Our detection network builds upon the backbone structure of S3FD \cite{zhang2017s3fd}. Each \rpool-Face model is created by placing \rpool Block directly after the input image or after a strided convolution (\rpool-Face-Quant). Following the \rpoollayer, we apply standard S3FD architecture for detection. Detection layers are placed at strides of 4, 8, 16, 32, 64, and 128, for square anchor boxes of sizes 16, 32, 64, 128, 256, and 512 as in S3FD. 

Following S3FD architecture, we fix the required receptive field size of each of the detection layers, which is then used to compute the number of MBConv Blocks or convolution layers after \rpool and before each detection layer. We also use S3FD's anchor matching strategy and the max-out background label technique. 

Images are trained on 640 $\times$ 640 images. A multi-task loss is used where cross-entropy loss is used for classification of anchor box and smooth L1 loss is used as regression loss for bounding box coordinate offsets. We use multi-scale testing and Non-Maximal Suppression during inference to determine final bounding boxes.

\begin{table}[t]
\centering
    \caption{\small The architecture of \rpool-Face-C}
\resizebox{0.5\columnwidth}{!}{
 \begin{tabular}{ c | c | c | c | c | c }
      \toprule
	  {Input} & {Operator} & $t$ & $c$ & $n$ & $s$\\
	  \midrule
	  $640 \times 480 \times 3$ & \rpoollayer & 1 & 64 & 1 & 4\\
	  $160 \times 120 \times 64$ & bottleneck & 6 & 24 & 2 & 1\\
	  $160 \times 120 \times 24$ & bottleneck & 6 & 32 & 3 & 2\\
	  $80 \times 60 \times 32$ & bottleneck & 6 & 64 & 4 & 2\\
	  $40 \times 30 \times 64$ & bottleneck & 6 & 96 & 3 & 2\\
	  $20 \times 15 \times 96$ & bottleneck & 6 & 160 & 2 & 2\\
	  $10 \times 7 \times 160$ & bottleneck & 6 & 320 & 1 & 2\\
	  \bottomrule
  \end{tabular}}
  \label{tab:rpool-face-c}

 \end{table}

Table~\ref{tab:rpool-face-c} contains the architecture of \rpool-Face-C. There is a detection layer after every bottleneck stack. The detection layer contains two $3 \times 3$ constitutional kernels which predict the class probability (2 outputs per pixel) and bounding box offsets(4 outputs per pixel). The convention followed in the table below is the same as in Table \ref{tab:mobilenetv2-rp}. t is the expansion coefficient, c is the number of output channels, n is the number of repetitions of the MBConv\footnote{We use the terms 'bottleneck', MBConv, and inverted residual interchangeably, they refer to the same block.} layer and s is the stride associated with the first of those stack of layers. \rpool's hidden state sizes are fixed to be: $h_1=h_2=16$.

\begin{table}[ht]
\centering
    \caption{\small The architecture of \rpool-Face-B}
\begin{tabular}{ c | c | c | c | c | c }
      \toprule
	  {Input} & {Operator} & $t$ & $c$ & $n$ & $s$\\
	  \midrule
	  $640 \times 480 \times 3$ & \rpoollayer & 1 & 24 & 1 & 4\\
	  $160 \times 120 \times 24$ & conv2d $3\times3$ & 1 & 24 & 4 & 1\\
	  $160 \times 120 \times 24$ & conv2d $3\times3$ & 1 & 96 & 1 & 2\\
	  $80 \times 60 \times 96$ & conv2d $1\times1$ & 1 & 32 & 1 & 1\\
	  $80 \times 60 \times 32$ & bottleneck & 6 & 32 & 3 & 1\\
	  $80 \times 60 \times 32$ & bottleneck & 6 & 64 & 3 & 2\\
	  $40 \times 30 \times 64$ & bottleneck & 6 & 128 & 2 & 2\\
	  $20 \times 15 \times 128$ & bottleneck & 6 & 160 & 1 & 2\\
	  $10 \times 7 \times 160$ & bottleneck & 6 & 320 & 1 & 2\\
	  \bottomrule
  \end{tabular}
  \label{tab:rpool-face-b}
  \end{table}

Architecture for \rpool-Face-B is shown in Table~\ref{tab:rpool-face-b}. The detection heads are after the second row of the table and then after each stack of bottleneck layers. \rpool's hidden state sizes are fixed to be: $h_1=h_2=6$.

\begin{table}
\vspace{-4mm}
\centering
    \caption{\small The architecture of \rpool-Face-A}
\resizebox{0.5\columnwidth}{!}{%
\begin{tabular}{ c | c | c | c | c | c }
      \toprule
	  {Input} & {Operator} & $t$ & $c$ & $n$ & $s$\\
	  \midrule
	  $640 \times 480 \times 3$ & \rpoollayer & 1 & 16 & 1 & 4\\
	  $160 \times 120 \times 16$ & Depthwise+Pointwise & 1 & 16 & 4 & 1\\
	  $160 \times 120 \times 16$ & Depthwise+Pointwise & 1 & 16 & 1 & 2\\
	  $80 \times 60 \times 16$ & bottleneck & 1 & 16 & 3 & 1\\
	  $80 \times 60 \times 16$ & bottleneck & 1 & 24 & 3 & 2\\
	  $40 \times 30 \times 24$ & bottleneck & 1 & 32 & 2 & 2\\
	  $20 \times 15 \times 32$ & bottleneck & 2 & 128 & 1 & 2\\
	  $10 \times 7 \times 128$ & bottleneck & 2 & 160 & 1 & 2\\
	  \bottomrule
  \end{tabular}}
 \label{tab:rpool-face-a}
 \vspace{-3mm}
 \end{table}
Architecture for \rpool-Face-A is shown in Table~\ref{tab:rpool-face-a}. The detection heads are after the second row of the table and then after each stack of bottleneck layers. \rpool's hidden state sizes are fixed to be: $h_1=h_2=16$. Depthwise+Pointwise refers to a depthwise separable $3 \times 3$ convolution followed by a pointwise $1 \times 1$ convolution.
                                                
\begin{table}[ht]
\begin{minipage}[b]{0.5\columnwidth}\centering
\caption{\small The architecture of \rpool-Face-Quant}
\resizebox{0.85\columnwidth}{!}{
\begin{tabular}{ c | c | c | c | c | c }
      \toprule
	  {Input} & {Operator} & $t$ & $c$ & $n$ & $s$\\
	  \midrule
	  $640 \times 480 \times 3$ & conv2d $3\times3$ & 1 & 4 & 1 & 2\\
	  $320 \times 240 \times 4$ & conv2d $3\times3$ & 1 & 4 & 1 & 1\\
	  $320 \times 240 \times 4$ & \rpoollayer & 1 & 32 & 1 & 4\\
	  $80 \times 60 \times 32$ & bottleneck & 2 & 16 & 4 & 1\\
	  $80 \times 60 \times 16$ & bottleneck & 2 & 24 & 4 & 2\\
	  $40 \times 30 \times 24$ & bottleneck & 2 & 32 & 2 & 2\\
	  $20 \times 15 \times 32$ & bottleneck & 2 & 64 & 1 & 2\\
	  $10 \times 7 \times 64$ & bottleneck & 2 & 96 & 1 & 2\\
	  \bottomrule
  \end{tabular}}
  \label{tab:rpool-quant}
\end{minipage}
\begin{minipage}[b]{0.5\columnwidth}
\centering
    \caption{\small The architecture of \rpool-Face-M4}
    \resizebox{0.85\columnwidth}{!}{
\begin{tabular}{ c | c | c | c | c | c }
      \toprule
	  {Input} & {Operator} & $t$ & $c$ & $n$ & $s$\\
	  \midrule
	  $320 \times 240 \times 1$ & conv2d $3\times3$ & 1 & 4 & 1 & 2\\
	  $160 \times 120 \times 4$ & \rpoollayer & 1 & 64 & 1 & 4\\
	  $40 \times 30 \times 64$ & bottleneck & 2 & 32 & 1 & 1\\
	  $40 \times 30 \times 32$ & bottleneck & 2 & 32 & 1 & 1\\
	  $40 \times 30 \times 32$ & bottleneck & 2 & 64 & 1 & 2\\
	  $20 \times 15 \times 64$ & bottleneck & 2 & 64 & 1 & 1\\
	  \bottomrule
  \end{tabular}}
 \label{tab:rpool-m4}
\end{minipage}
\end{table}
The architecture for \rpool-Face-Quant is shown in Table~\ref{tab:rpool-quant}. The detection heads are after the second row of the table and then after each stack of bottleneck layers. The first detection head has a strided $3\times3$ convolution to reach a total stride of 4 (following S3FD). \rpool's hidden state sizes are fixed to be:  $h_1=h_2=4$.

Table~\ref{tab:rpool-m4} shows the \rpool-Face-M4 architecture for our cheapest model deployed on a M4 device. The model has 4 detection layers after each MBConv Block. \rpool's hidden state sizes are fixed to be: $h_1=h_2=16$.

\begin{figure*}[h]
	\includegraphics[width=\textwidth]{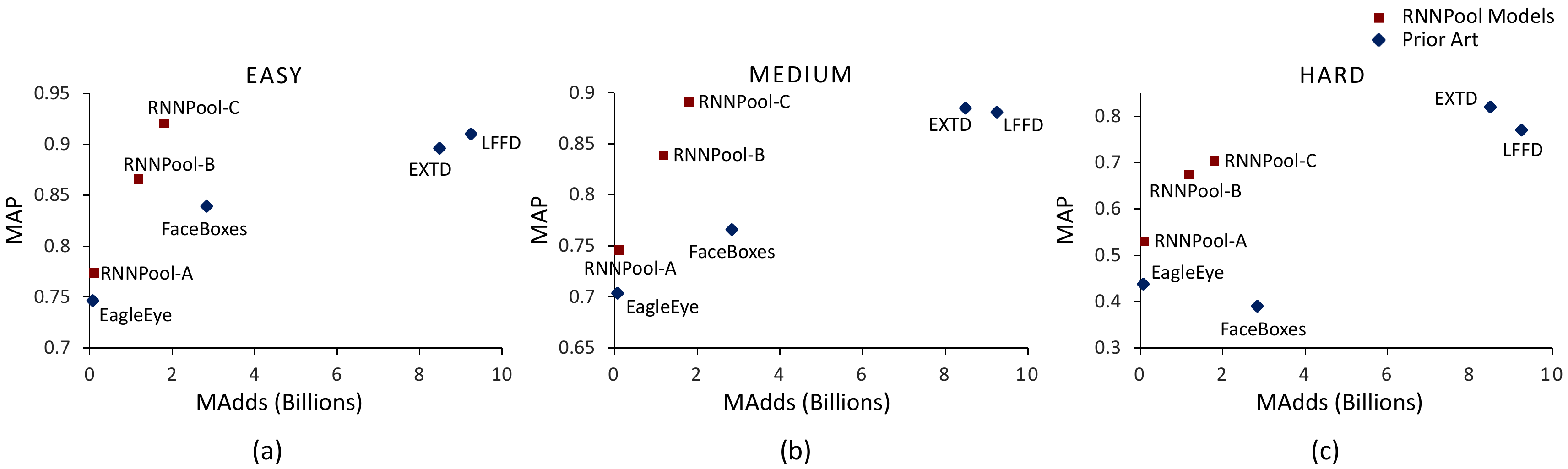}
	\caption{WIDER Face Dataset: MAdds  vs MAP of various methods including \rpool+S3FD.}
	\label{fig:facedetection}
	\vspace{-5mm}
\end{figure*}

The RNNPool models decrease MAdds drastically while maintaining performance. Figure \ref{fig:facedetection}, shows the difference we are making. When restricted to the methods with $<$2G MAdds requirement, our model attains even better MAP (for easy and medium dataset) than the state-of-the-art EXTD and LFFD architectures (which need about  10G MAdds per inference.

\section{Hyperparameters}
\label{sec:hyperparams}
Models are trained in PyTorch~\citep{paszke2019pytorch} using SGD with
momentum optimizer~\citep{sutskever2013importance} with weight decay
$4\times 10^{-5}$ and momentum $ 0.9$.  We do data-parallel training
with 4 NVIDIA P40 GPUs and use a batch size of 256 for classification
and 32 for face detection.  We use a cosine learning rate schedule
with an initial learning rate of $0.05$ for classification tasks, and
$0.01$ with 5 warmup epochs for face detection tasks. All convolution
layers use learnable batch normalization. We use the
EdgeML~\citep{edgeml04} implementation of FastGRNN. All ImageNet-10 and face detection
experiments were trained for 300 epochs. Both Visual Wake Words and ImageNet-1K experiments were run for 150 epochs. Best top-1 validation accuracy is reported in all the classification datasets and test MAP was reported for face detection. 

We use FastGRNN as both the RNNs in \rpool. We usually use the same hidden
dimension for both the RNNs. We fix $\zeta$ as 1 and $\nu$ as 0 for
all models, for stability, and use piecewise linear non-linearities
quantTanh and quantSigmoid for the Visual Wake Word models, so we can
quantize it without loss of information.

Various image augmentations were used for training each network. For
the ImageNet experiments, the training images were cropped to a random
size of 0.08 to 1.0 times the original size and reshaped to a random
aspect ratio of 3/4 to 4/3. This was then resized to 224 $\times$
224. This image was further flipped horizontally randomly and then
normalized by the mean and standard deviation. For the validation set, we
resize the input image to 256 $\times$ 256 and then take a center crop
of 224 $\times$ 224. For the Visual Wake Word experiment, we follow a
similar process except during training we crop the input image first
to a random size of 0.2 to 1.0 times the original size. For varying
resolutions from 96 to 224 as reported in Figure \ref{fig:vww}, the
ratio of resizing resolution of the input image and center crop size is kept
the same during validation. All other augmentations are kept the same with
output size changed from 96 to 224. For Face Detection experiments we
use augmentations like in S3FD~\citep{zhang2017s3fd}. This includes
color distortion, random cropping: specifically zooming in to
smaller faces to get larger faces to train on, and horizontal
flipping after cropping to 640 $\times$ 640. Note that the same
augmentation strategies were used for the baseline models also for
a fair comparison.

\section{\rpool Ablation}
\label{sec:ablation}
In this section, we first discuss the changes in accuracy, peak RAM, MAdds, and the number of parameters on varying hyperparameters of \rpool like patch size, hidden dimensions, and stride. We also compare the same for multiple layers of \rpool. We use MobileNetV2 as the base network and the dataset is ImageNet-10. Note that the first row refers to the MobileNetV2-\rpool architecture in Table~\ref{tab:mobilenetv2-rp}, and the other rows (b)-(e) of Table~\ref{tab:ablation} are variations on it. Table~\ref{tab:ablation} (f) and (g) have another 4 MBConv blocks replaced in the MobileNetV2-\rpool architecture (Row 3 of Table \ref{tab:mobilenetv2-rp}). (f) uses a single \rpool to do this replacement whereas (g) uses two consecutive \rpool Blocks. All variations have $\sim$2M parameters (even (g) which has 2 \rpool layers has a very minimal model size overhead). This suggests that a finer hyperparameter and architecture search could lead to a better trade-off between accuracy and compute requirements.
\begin{table}[h]
\centering
    \caption{\small Comparison of accuracy, peak RAM and MAdds for variations in hidden dimensions, patch size and stride in RNNPool for MobileNetV2 and on ImageNet-10 dataset. Parameters are same as the base if not mentioned. (f) and (g) are further replacements in MobileNetV2-\rpool (Row 3 of Table~\ref{tab:mobilenetv2-rp}).}
\resizebox{1\columnwidth}{!}{%
\begin{tabular}{c c c c c c}
  \toprule
  \# & Hyperparameters  & Accuracy (\%) &  Peak RAM &  MAdds\\
  \midrule
  (a) & Reported (Patch Size = 6; $h_1=h_2=16$, Stride = 4) & 94.4 & 0.24MB & 0.23G\\  
  (b) & Patch size = 8 & 94.0 & 0.24MB & 0.24G\\  
  (c) & Patch size = 4 & 93.2 & 0.24MB & 0.22G\\  
  (d) & $h_1=h_2=8$ & 92.8 & 0.14MB & 0.21G\\
  (e) & $h_1=h_2=32$ & 95.0 & 0.43MB & 0.29G\\
  (f) & Stride = 8; Patch Size = 12 & 94.0 & 0.14MB & 0.17G\\
  (g) & Stride = 4; Patch Size = 6 and Stride = 2; Patch Size = 4 & 93.2 & 0.19MB & 0.17G\\
  \bottomrule
\end{tabular}}
\label{tab:ablation}
\end{table}

In Table~\ref{tab:otherrnn}, we ablate over the choice of RNN cell (LSTM, GRU and FastGRNN) in \rpool for the MobileNetV2-\rpool model (Table~\ref{tab:mobilenetv2-rp}) on the ImageNet-10 dataset. We show that the choice of FastGRNN results in significantly lower MAdds than LSTM or GRU while having about 1\% higher accuracy. Finally, Table~\ref{fig:log} has the training curve for the MobileNetV2-\rpool on ImagetNet-10 showing that training with \rpool is not harder than the base models.

\begin{table}[h!]
  \centering
  \begin{minipage}[t]{0.45\columnwidth}
    \includegraphics[width=\columnwidth]{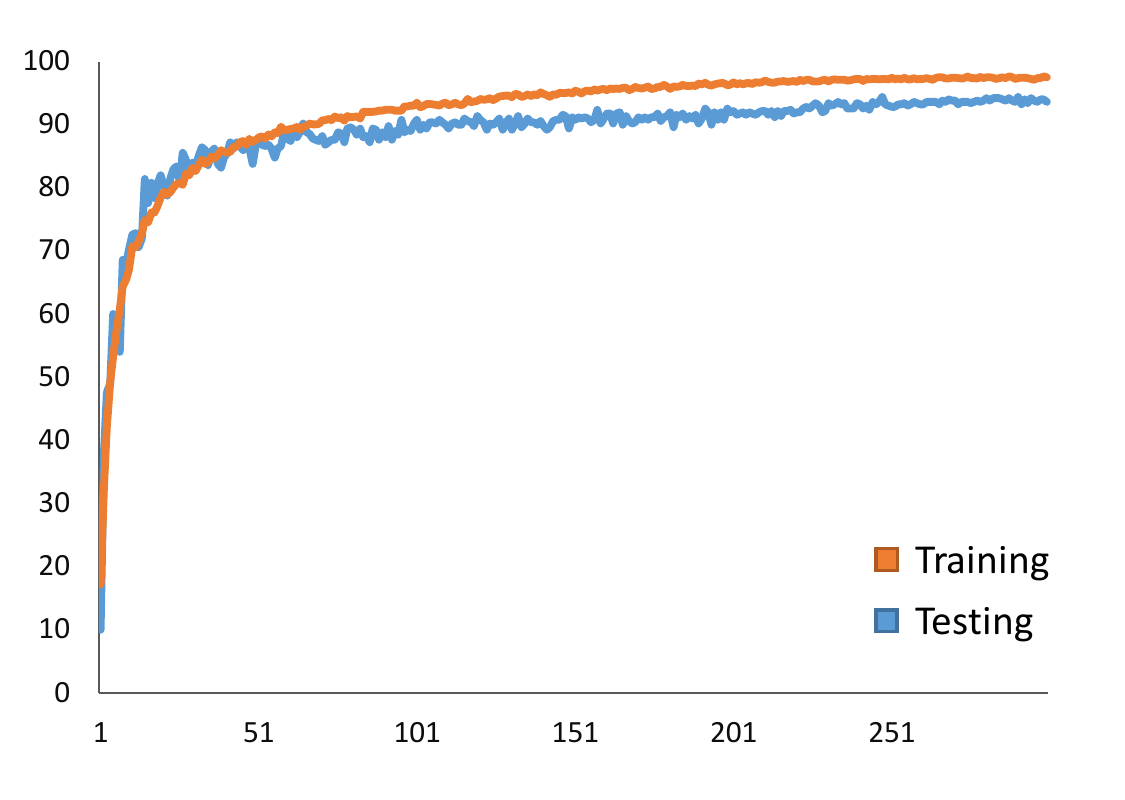}
    \vspace{-18pt}
    \caption{\small Training curve of MobileNetV2-\rpool on ImageNet-10.}
    \label{fig:log}
  \end{minipage}
  \hfill
  ~
  \begin{minipage}[t]{0.50\columnwidth}
  \centering
  \scriptsize
    \vspace{-78pt}
    \caption{\small Ablation over RNN cell in \rpool for MobileNetV2-\rpool on ImageNet-10.}
    \resizebox{\columnwidth}{!}{%
    \begin{tabular}{l c c c c}
      \toprule
      \multicolumn{1}{l}{RNN cell} & {Parameters} & {MAdds} & {Accuracy (\%)} \\
      \midrule
      LSTM & 2.0M & 266M & 93.4\\
      GRU & 2.0M & 246M & 93.0\\
      FastGRNN & \textbf{2.0M} & \textbf{226M} & \textbf{94.4} \\\bottomrule
    \end{tabular}}
    \label{tab:otherrnn}
  \end{minipage}
  \vspace{-17pt}
\end{table}
\section{Face Detection Qualitative Results}
\label{sec:facedetviz}
\begin{figure}[h!]
\begin{minipage}{.5\columnwidth}
\centering
	\includegraphics[width=1\columnwidth]{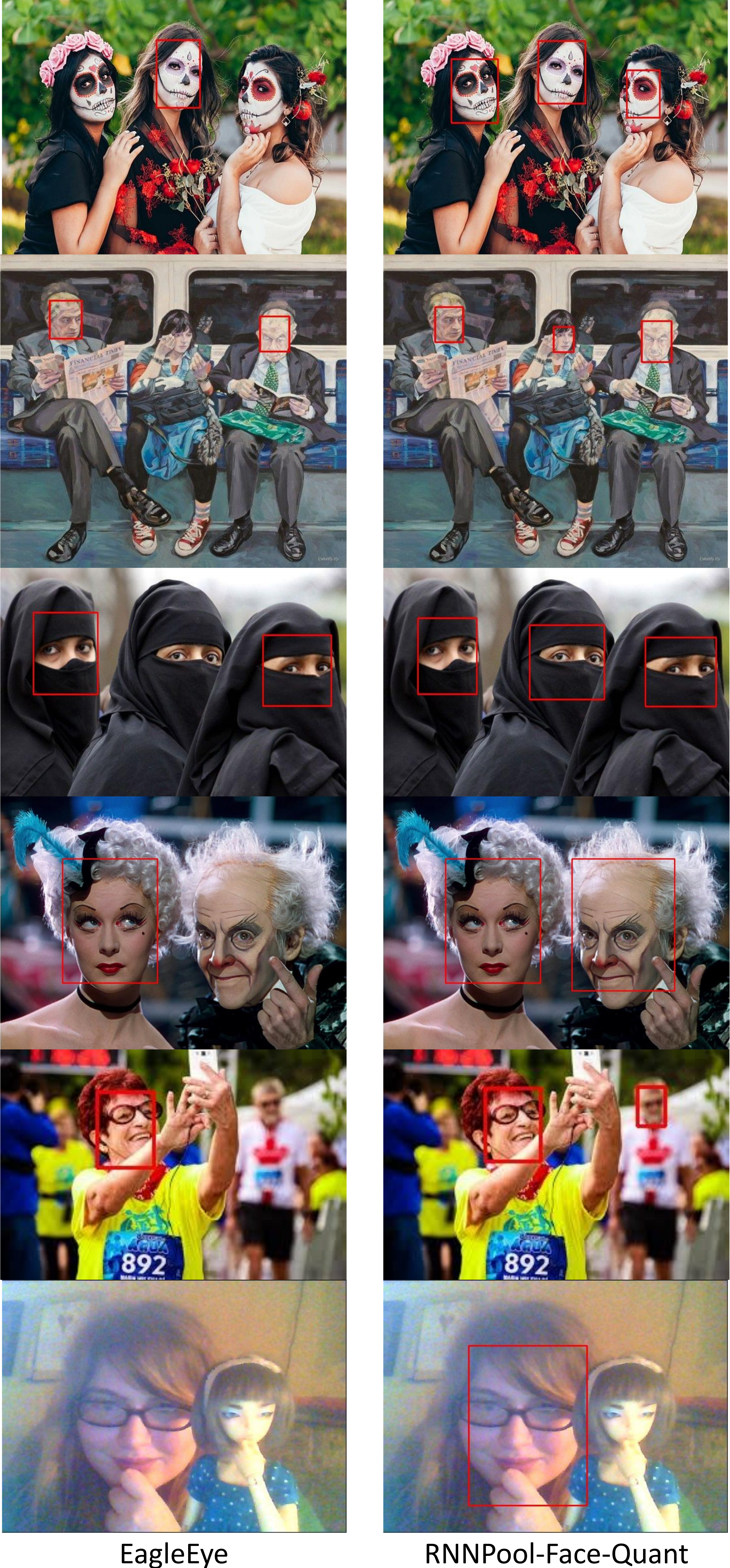}
	\caption{\small Comparison of performance on test images with
          Eagle-Eye and RNNPool-Face-Quant. The confidence threshold
          is set to 0.6 for both models. EagleEye misses faces when
          there is makeup, occlusion, blurriness and in grainy
          pictures, while our method detects them. However, in the
          case of some hard faces, RNNPool-Face-Quant misses a few of
          them or does not draw a bounding box over the full face.}
	\label{fig:faceviz}
	\end{minipage}
	\quad
\begin{minipage}{.5\columnwidth}
\centering
	\includegraphics[width=1\columnwidth]{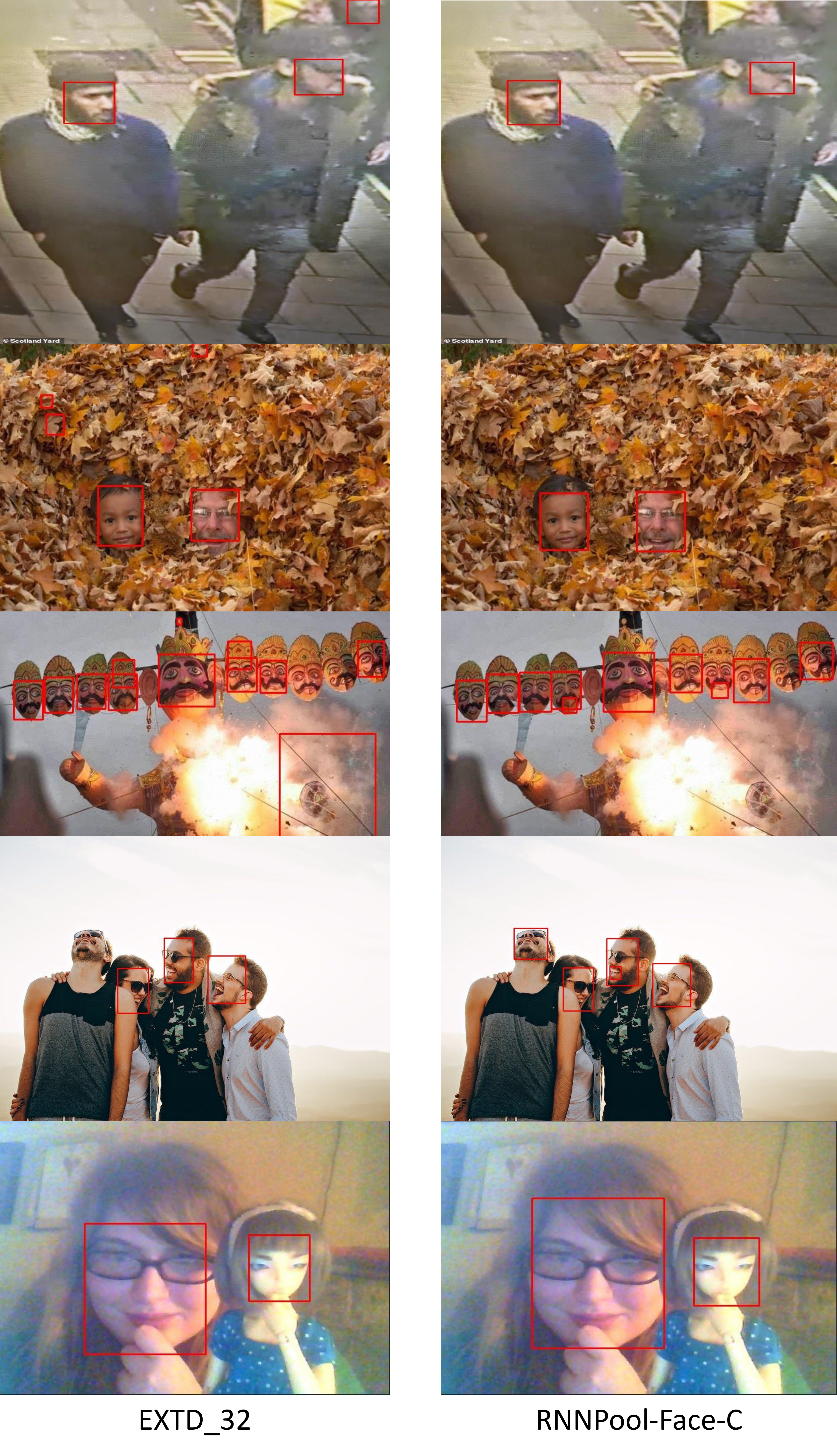}
	\caption{\small Comparison of performance on test images with
          EXTD\_32 and RNNPool-Face-C. The confidence threshold is set
          to 0.6 for both models. The EXTD model has more false
          positives and misses more faces. In the first image, EXTD
          makes a faulty prediction at the top right. In the second
          image, EXTD mistakes regions in leaves for faces, while our
          model detects just the two correct faces. In the next image,
          both the models have some wrong detections, but the EXTD
          model detects a large bounding box that is a false
          positive. In the next image EXTD misses a face with an
          unnatural pose that our model detects. However, our model
          detects a face within a face which in general can be removed
          easily. In the next image (last row above), both the models
          detect the two faces, which weren't detected by the models
          on the left. Our model detects a slightly better bounding
          box than EXTD.  }
	\label{fig:faceviz_big}
		\end{minipage}
\end{figure}

Figures~\ref{fig:faceviz} and~\ref{fig:faceviz_big} show the
qualitative results where \rpool based models outperform the current
state-of-the-art real-time face detection models.

\end{document}